%% file: neurips_2026.tex
\documentclass{article}

\usepackage[preprint]{neurips_2026}

\usepackage[utf8]{inputenc} 
\usepackage[T1]{fontenc}    

\usepackage{amsmath,amssymb,amsfonts}
\usepackage{mathrsfs}
\usepackage{nicefrac}

\usepackage{booktabs}
\usepackage{multirow}
\usepackage{array}
\usepackage{colortbl}

\usepackage{graphicx}
\usepackage{wrapfig}
\usepackage{subfigure}

\usepackage{algorithm}
\usepackage{algorithmic}

\usepackage{soul}
\usepackage{pifont}
\usepackage{textcomp}
\usepackage{xcolor}

\usepackage{url}
\usepackage[hidelinks]{hyperref}

\usepackage{microtype}

\title{DisCo-FLoc: Semantic-Free Floorplan Localization via $SE(2)$-Aware Contrastive Disambiguation}

\author{
Ping Zhong$^{1}$ \quad
Shiyong Meng$^{1}$ \quad
Bolei Chen$^{1}$\thanks{Corresponding author: \texttt{boleichen@csu.edu.cn}.} \quad
Tao Zou$^{1}$ \quad
Chaoxu Mu$^{2}$ \quad
Jianxin Wang$^{1}$ \\
$^{1}$School of Computer Science and Engineering, Central South University \\
$^{2}$School of Electrical and Information Engineering, Tianjin University \\
}

\begin{document}

\maketitle

\begin{abstract}
Visual \textbf{F}loorplan \textbf{Loc}alization (FLoc) struggles with severe structural aliasing caused by repetitive minimalist layouts. This occurs because physically distant poses share highly similar visual-geometric features, which degrades spatial separability and angular discriminability. While existing methods attempt to mitigate these ambiguities by relying on costly semantic annotations, the resulting performance gains remain inherently limited. To address the above issues, we propose \textcolor{blue}{\textbf{DisCo}}-FLoc, a semantic-free method for visual-geometric \textcolor{blue}{\textbf{Co}}ntrastive \textcolor{blue}{\textbf{Dis}}ambiguation. First, we introduce a depth-aware \textbf{R}ay \textbf{R}egression \textbf{P}redictor (RRP) that serves as a dense-to-ray geometric projector. By explicitly suppressing visual clutter along the vertical dimension, RRP projects monocular RGB images into 2D ray primitives, which are matched with floorplans to produce geometry-aware FLoc candidates. Second, to resolve the remaining ambiguity among these candidates, we propose a spatially perturbed contrastive objective to align RGB images with local floorplan structures and formulate a visual-geometric compatibility function. In particular, we meticulously construct positive and negative samples at both positional and directional levels through $SE(2)$ pose perturbations for contrastive learning, effectively achieving pose smoothness, spatial separability, and angular discriminability. The compatibility function enables DisCo-FLoc to disambiguate FLoc by using richer visual context beyond pure geometric layouts, without requiring any semantic annotations. Extensive experiments on two challenging visual FLoc benchmarks demonstrate that DisCo-FLoc significantly outperforms state-of-the-art semantic-based methods, especially narrowing the performance gap between positional and directional FLoc accuracy.
\end{abstract}

\section{Introduction}


\begin{figure}[!t]
 \centering
 \includegraphics[width=1.0\linewidth]{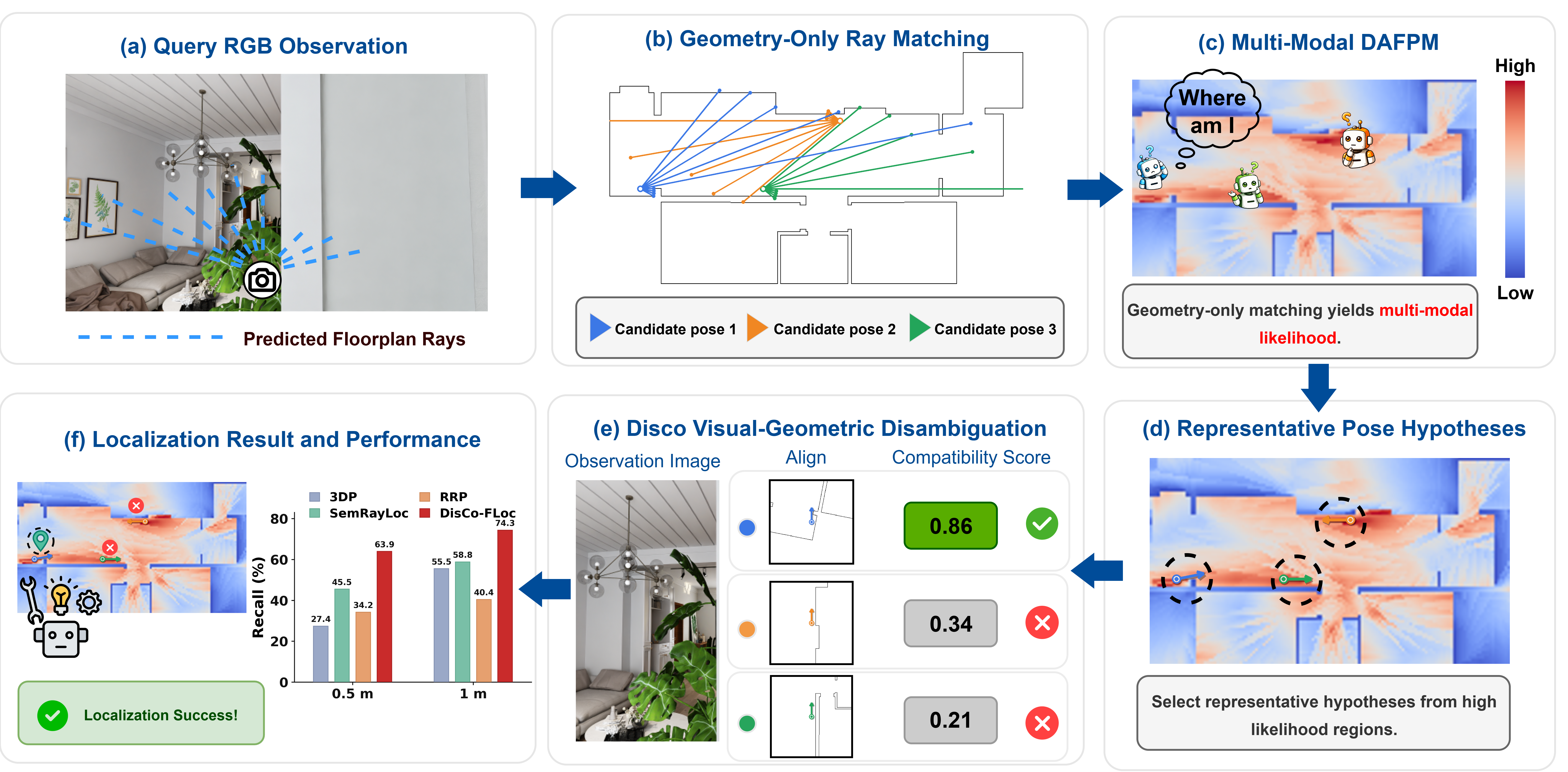}
 \vspace{-0.5cm}
 \caption{Due to modal gap and inherent structural aliasing, geometry-only ray matching can easily degrade spatial separability and angular discriminability (a)-(b), thus resulting in multi-modal FLoc probability distributions (c). To address this challenge, we perform visual-geometric disambiguation (e) on the pose hypotheses refined from DAFPM (d). Our method significantly outperforms existing semantic-based methods without using any semantic annotations (f).}
 \label{fig1}
 \vspace{-0.7cm}
\end{figure}

Camera localization is a fundamental task in computer vision, serving as a prerequisite for various applications including robotics \cite{li2024flona,huang2025floor} and augmented reality. Visual localization in indoor environments remains challenging due to complex layouts and the unavailability of satellite signals. Conventional methods typically rely on pre-built 3D models \cite{liu2017efficient,sarlin2019coarse,sattler2016efficient} or large-scale image databases \cite{balntas2018relocnet,2017NetVLAD}. However, these approaches are storage-intensive and require frequent maintenance, which limits their scalability. In contrast, building floorplans provide a persistent and accessible geometric reference that is invariant to appearance changes, such as lighting variations or furniture rearrangements. Consequently, visual \textbf{F}loorplan \textbf{L}ocalization (FLoc) \cite{chen2024f3loc} has emerged as a practical alternative for long-term indoor localization.

Visual FLoc is primarily confronted by two critical challenges. The \textit{first} is the cross-modal gap between high-dimensional RGB observations and minimalist 2D floorplans, as shown in Fig.\ref{fig1} (a) and (b). While monocular images contain complex scene information such as furniture, varying illumination, and visual occlusions, floorplans preserve only fundamental geometric layouts. Although existing methods attempt to bridge this gap by explicitly matching 2D geometric primitives \cite{karkus2018particle,chen2024f3loc} or implicitly learning 3D scene priors \cite{chen20253dp,chen2025perspective}, they either yield only limited performance gains or introduce computationally expensive pre-training. 

The \textit{second} challenge is structural aliasing inherent in repetitive indoor environments, which significantly degrades spatial separability and angular discriminability. In this context, spatial separability refers to the capacity to distinguish between physically distant locations that share near-identical geometric profiles. A lack of such separability typically results in multi-modal probability distributions, as shown in Fig.\ref{fig1} (c). Angular discriminability denotes the sensitivity required to identify the correct heading among orientations with similar observation signatures, particularly in symmetric spaces. To mitigate these ambiguities, conventional approaches \cite{min2022laser,grader2025supercharging} rely on auxiliary semantic labels like doors or windows to introduce distinctive landmarks. However, the efficacy of these methods is constrained by the high cost and limited availability of semantic annotations. Furthermore, high-level semantics often fail to capture the fine-grained visual-geometric nuances necessary for robust $SE(2)$ disambiguation when the semantic categories themselves are distributed repetitively.

To address the above challenges, we propose \textcolor{blue}{\textbf{DisCo}}-FLoc, a semantic-free method for visual-geometric \textcolor{blue}{\textbf{Co}}ntrastive \textcolor{blue}{\textbf{Dis}}ambiguation. First, we introduce a depth-aware \textbf{R}ay \textbf{R}egression \textbf{P}redictor (RRP) as a dense-to-ray geometric projector. RRP suppresses visual clutter along the vertical dimension and transforms monocular RGB observations into 2D ray primitives. These predicted rays are matched against floorplan geometry to generate a \textbf{D}epth-\textbf{A}ware \textbf{F}Loc \textbf{P}robabilistic \textbf{M}ap (DAFPM), providing an initial set of geometry-aware FLoc candidates, as shown in Fig.\ref{fig1} (c). While this stage reduces the visual-floorplan modal gap, repetitive layouts may still yield multiple plausible candidates.

Second, to resolve residual ambiguities among the generated candidates, we propose a spatially perturbed contrastive objective to formulate a visual-geometric compatibility function. This compatibility function enables DisCo-FLoc to disambiguate FLoc by using richer visual context beyond pure geometric layouts, without requiring any semantic annotations. In practice, we first refine the multi-modal DAFPM to select representative pose hypotheses, as shown in Fig.\ref{fig1} (d). Then, the compatibility function aligns RGB observations with local floorplan structures based on visual-geometric correspondences (as shown in Fig.\ref{fig1} (e)), thereby generating a \textbf{D}isambiguation \textbf{P}robability \textbf{M}ap (DPM) formed by compatibility scores. Such visual-geometric correspondences are pre-modeled through $SE(2)$-aware contrastive learning. Technically, slack positives correlate images with floorplan geometries under minor pose perturbations to promote local pose smoothness. Simultaneously, position-level negatives suppress physically distant but visually similar candidates, while orientation-level negatives enforce sensitivity to heading changes at the same location.

Extensive evaluations on two challenging benchmarks demonstrate that DisCo-FLoc achieves \textbf{S}tate-\textbf{O}f-\textbf{T}he-\textbf{A}rt (SOTA) performance without using any semantic annotations, significantly outperforming existing semantic-based methods, as shown in Fig.\ref{fig1} (f). Our main contributions are as follows:

(1) We propose DisCo-FLoc, a semantic-free method that integrates depth-aware candidate generation with $SE(2)$-aware contrastive learning to mitigate structural aliasing in visual FLoc.

(2) We introduce a depth-aware RRP acting as a dense-to-ray geometric projector. By suppressing vertical visual clutter and transforming monocular RGB observations into 2D ray primitives, RRP generates geometry-aware pose candidates to bridge the visual-floorplan modal gap.

(3) We design a spatially perturbed contrastive objective to align RGB observations with local floorplan structures. By incorporating $SE(2)$-aware constraints via slack positives and dual-level negatives, this objective enhances local pose smoothness, spatial separability, and angular discriminability without requiring semantic annotations.

\vspace{-0.3cm}
\section{Related Work}

\textbf{Visual Localization.} Visual localization is a fundamental problem in computer vision and has been widely studied. Traditional methods include image retrieval techniques \cite{balntas2018relocnet,2017NetVLAD}, which find the most similar images in a database and estimate the pose of the query image based on the retrieved ones. Structure-from-motion-based approaches \cite{panek2022meshloc,sarlin2019coarse} build a 3D model of the environment and establish 2D-3D correspondences by matching local descriptors, computing camera poses using minimal solvers \cite{kukelova2008automatic} and RANSAC \cite{fischler1981random} or its recent variants \cite{barath2021graph}. Scene coordinate regression methods \cite{brachmann2017dsac} learn to regress the 3D coordinates of image pixels, while pose regression techniques \cite{kendall2017geometric} use networks to directly predict a 6-DoF camera pose from input images. These methods often rely on pre-built 3D models that are storage-intensive and scene-specific, limiting their applicability in unseen environments. The recently emerging visual FLoc task provides a promising alternative by leveraging readily available 2D floorplans.

\textbf{Visual Floorplan Localization.} Visual FLoc is conceptually related to LiDAR-based \textbf{M}onte \textbf{C}arlo \textbf{L}ocalization (MCL) \cite{dellaert1999monte,chu2015you,mendez2018sedar,winterhalter2015accurate}. While MCL is a classic framework for 2D localization on geometric maps, its reliance on LiDAR sensors limits deployment on common mobile devices. To address this limitation, various studies \cite{boniardi2019robot,chu2015you,howard2022lalaloc++,howard2021lalaloc,min2022laser} have explored visual FLoc using monocular or panoramic images. These methods typically involve matching 2D scene priors \cite{boniardi2019robot} or visual features \cite{min2022laser} with scene layouts. Other approaches compare query images against synthetic panoramic features rendered from the floorplan \cite{howard2022lalaloc++,howard2021lalaloc}. However, many of these techniques assume prior knowledge of camera and room heights or require panoramic inputs, which constrains their applicability in general monocular settings.

More recently, generic monocular visual FLoc techniques \cite{karkus2018particle,chen2024f3loc,chen20253dp} have utilized Bayesian filters \cite{jonschkowski2016end,bishop2001introduction} for sequential localization. Despite their progress, these methods remain susceptible to structural aliasing in repetitive minimalist environments. Physically distant poses can generate near-identical visual-geometric responses, which often causes multi-modal uncertainty and ambiguous pose candidates. To mitigate these ambiguities, several methods \cite{min2022laser,mendez2020sedar,grader2025supercharging} incorporate sparse semantic labels such as room categories, doors, or windows as auxiliary cues to distinguish similar geometric structures. While recent SOTA method \cite{chen2025perspective} reduces annotation costs by learning unsupervised \textbf{multi-modal alignments}, it primarily relies on costly 3D cues. Such methods do not explicitly enhance the spatial separability and angular discriminability of visual-geometric representations.

Unlike existing methods that rely on costly semantic labels or 3D cues to resolve ambiguities, DisCo-FLoc first generates geometry-aware pose candidates through a depth-aware dense-to-ray projection mechanism. Then, it employs $SE(2)$-aware contrastive constraints to enhance spatial separability and angular discriminability, achieving effective FLoc disambiguation without using any semantic labels or additional data.

\section{Methodology}



\vspace{-0.3cm}
\subsection{Problem Formulation} \label{3.1}

This work formulates visual FLoc as estimating a camera pose within the $SE(2)$ space by aligning a monocular RGB observation with a 2D floorplan. Let $F \in \mathbb{R}^{\mathcal{H}\times \mathcal{W}}$ denote a floorplan that preserves geometric occupancy information without semantic categories. Given an RGB image $\mathcal{I}$, our objective is to estimate the absolute camera pose $S_{\mathcal{I},F}=(x,y,\theta)$, where $(x,y)\in \mathbb{R}^2$ represents the 2D spatial location and $\theta \in SO(2)$ represents the viewing orientation. Based on the joint observation $O_{\mathcal{I},F}=(\mathcal{I},F)$, we model FLoc probabilistically by estimating the conditional distribution $p(S_{\mathcal{I},F}|O_{\mathcal{I},F})$. We discretize the continuous $SE(2)$ pose space into a finite candidate set $\mathcal{S}=\{S_i\}$ and construct a probabilistic volume $P \in \mathbb{R}^{\hat{\mathcal{H}}\times \hat{\mathcal{W}}\times O}$. Each element $P(S_i)$ indicates the likelihood of a candidate pose $S_i$. Here, $\hat{\mathcal{H}}$ and $\hat{\mathcal{W}}$ denote the discretized spatial grid dimensions, and $O$ represents the number of orientation bins. In repetitive indoor environments, structural aliasing often assigns high likelihoods to physically distant or orientation-inconsistent candidates. This lack of spatial separability and angular discriminability inherently leads to a multi-modal probability volume. Our goal is to formulate a $SE(2)$-aware visual-geometric compatibility function that suppresses these ambiguous candidates. This function effectively enhances pose-structured separability and enables reliable maximum posteriori estimation:
\setlength\abovedisplayskip{0.15cm}
\setlength\belowdisplayskip{0.15cm}
\begin{flalign} \small
\begin{aligned}
\label{eq1}
\hat{S}_{\mathcal{I},F} = \arg\max_{S_i \in \mathcal{S}} p(S_i|O_{\mathcal{I},F}).
\end{aligned}
\end{flalign}

\vspace{-0.3cm}
\subsection{Stage 1: Dense-to-Ray Projection for Geometry-Aware Candidate Generation} \label{3.2}

Vision-to-floorplan alignment requires extracting compact geometric cues from cluttered RGB observations to match against 2D floorplan structures. Existing visual FLoc methods \cite{chen2024f3loc,chen20253dp,chen2025perspective} typically rely on observation models pre-trained for image classification \cite{deng2009imagenet} or generic contrastive learning \cite{xia2018gibson}. These models lack explicit geometric awareness and often fail to distinguish true room layouts from 3D object occlusions.

To bridge this visual-floorplan modal gap, we introduce a depth-aware RRP acting as a dense-to-ray geometric projector. Recognizing 2D ray-casting as a specialized form of depth estimation, we utilize a pre-trained depth encoder with a DINOv2-S backbone \cite{oquab2023dinov2,yang2024depth} to extract dense visual features. The RRP then employs a specialized attention mechanism to derive 2D layout primitives from these features. Specifically, the dense features pass through a convolutional layer to generate keys and values. A vertically weighted aggregation is subsequently applied to form 1D queries. This operation explicitly suppresses vertical visual clutter and isolates horizontal layout structures, effectively minimizing the impact of 3D occlusions. For spatial awareness, query positional encodings are derived from 1D coordinates, while key and value encodings are mapped from 2D image coordinates.

For each query, attention is applied across the image to predict the distance to the nearest wall. The depth prediction is formulated as a weighted sum of discrete bins:
\begin{flalign} \small
\begin{aligned}
\label{eq2}
d_i = \sum_{k=1}^{D} P_{i,k} \, d_k,\quad
d_k = \left(d_{\text{min}}^\gamma + \frac{k}{D}(d_{\text{max}}^\gamma - d_{\text{min}}^\gamma)\right)^{1/\gamma},
\end{aligned}
\end{flalign}
where $P_i \in \mathbb{R}^D$ denotes the predicted probability distribution across $D$ bins for the $i$-th ray, ensuring $\sum_{k=1}^{D} P_{i,k} = 1$ via a softmax function. The parameter $N$ ($i=1,\ldots,N$) denotes the total number of predicted rays. The terms $d_{\text{max}}$ and $d_{\text{min}}$ define the depth boundaries, while the power-law discretization parameter $\gamma$ controls the resolution distribution. The RRP is optimized using a joint $L_1$ and cosine similarity-based shape loss:
\begin{flalign} \small
\begin{aligned}
\label{eq3}
\mathcal{L}_{FLoc} = ||\mathbf{d} - \mathbf{d}^{*}||_1 + 1 - \frac{\mathbf{d}^{\top}\mathbf{d}^{*}}{\max\{||\mathbf{d}||_2||\mathbf{d}^{*}||_2,\epsilon \}},
\end{aligned}
\end{flalign}
where $\mathbf{d}$ and $\mathbf{d}^{*}$ represent the predicted and ground-truth 2D-ray depths respectively, and $\epsilon$ is a small constant to prevent division by zero.

During inference, the predicted 2D ray primitives are matched against ground-truth rays generated from the floorplan geometry \cite{chen2024f3loc} across all discrete candidate poses $S_i=(x,y,\theta)$. This geometric matching yields a DAFPM, denoted as $P_d \in [0,1]^{\hat{\mathcal{H}}\times\hat{\mathcal{W}}\times O}$. Each element $P_d(S_i)$ indicates the initial geometry-aware likelihood of a pose candidate $S_i$.

\textbf{Transition to Disambiguation.} Although the RRP bridges the modal gap and provides geometry-aware candidates, pure geometric matching remains susceptible to structural aliasing in repetitive floorplans. Consequently, the DAFPM often exhibits a multi-modal distribution with multiple peaks corresponding to physically distant or orientation-inconsistent poses. To resolve these residual ambiguities, we refine the Top-$K$ candidate poses extracted from the DAFPM and propose the spatially perturbed contrastive objective for FLoc disambiguation detailed in Sec. \ref{3.3}.

\begin{figure*}[!t]
 \centering
 \includegraphics[width=1.0\linewidth]{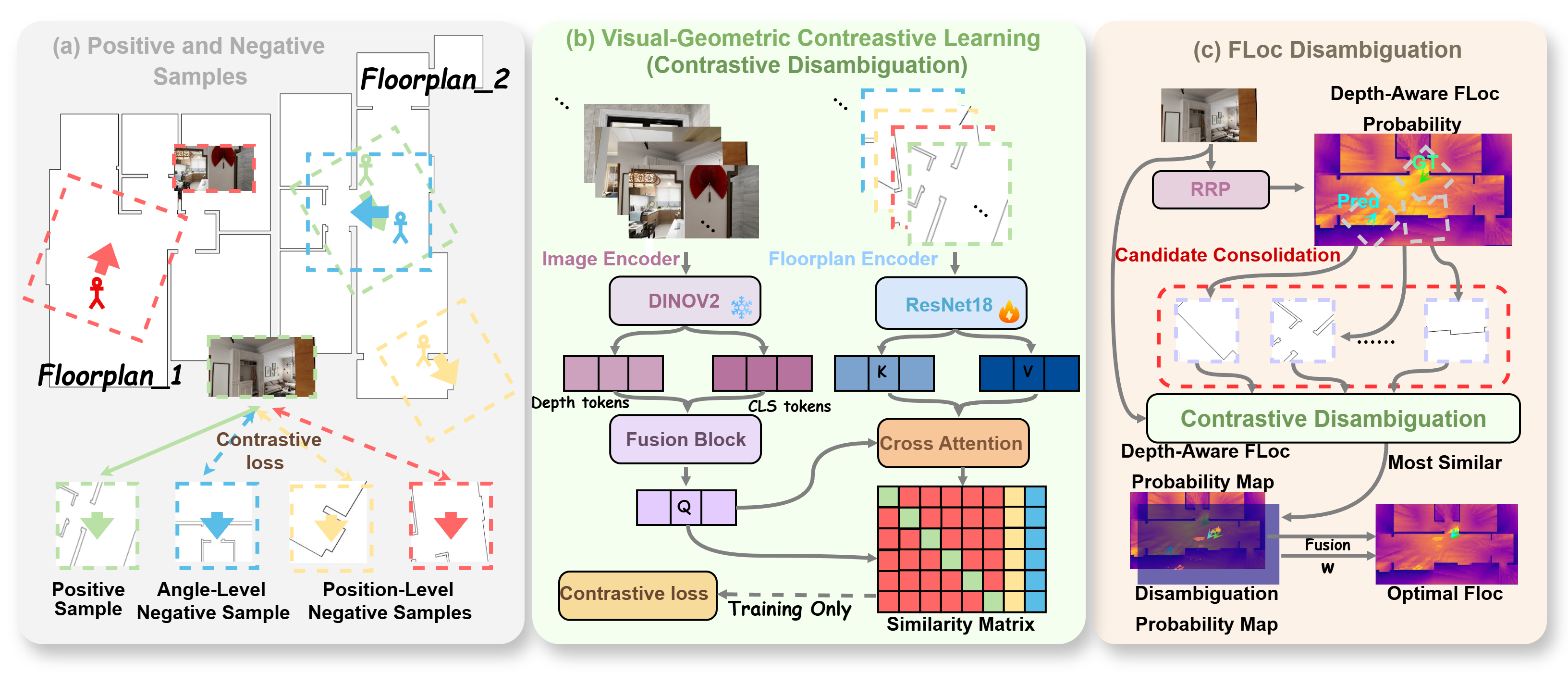}
 \vspace{-0.5cm}
 \caption{
 (a) illustrates the spatially perturbed sampling strategy, which constructs slack positives, position-level negatives, and orientation-level negatives according to the $SE(2)$ pose structure.
 (b) details the proposed visual-geometric contrastive disambiguation module.
 (c) shows the inference process: the DAFPM generated by RRP provides initial pose candidates, which are refined by the learned DPM. Fusing DAFPM with DPM suppresses ambiguous candidates and yields the final FLoc prediction.
 }
 \label{fig2}
 \vspace{-0.5cm}
\end{figure*}

\vspace{-0.3cm}
\subsection{Stage 2: $SE(2)$-Aware Contrastive Disambiguation} \label{3.3}

\textbf{Visual-Geometric Compatibility Modeling.}
Although Stage 1 provides geometry-aware pose candidates, repetitive floorplan structures often induce structural aliasing. This phenomenon manifests as multiple plausible peaks in the DAFPM. To resolve these residual ambiguities, we formulate a $SE(2)$-aware visual-geometric compatibility function between the query image and local floorplan structures cropped at candidate poses. Let $\mathbf{f}_j$ denote the visual feature of the $j$-th query image, and let $\mathbf{g}_l$ denote the geometric feature cropped from the floorplan under a candidate pose $S_l=(x_l,y_l,\theta_l)$. Their compatibility is defined by cosine similarity:
\begin{flalign} \small
\begin{aligned}
\label{eq4}
c(\mathcal{I}_j,F,S_l)=\operatorname{sim}(\mathbf{f}_j,\mathbf{g}_l).
\end{aligned}
\end{flalign}
We construct contrastive samples according to the geometry of the $SE(2)$ pose space rather than treating all candidate poses uniformly, as shown in Fig.\ref{fig2} (a). This structured constraint ensures that the learned compatibility maintains local pose smoothness, enhances spatial separability, and improves angular discriminability.

\textbf{Slack Positives for Local Pose Smoothness.}
For a query image $\mathcal{I}_j$ with a ground-truth pose $S_j=(x_j,y_j,\theta_j)$, we construct a slack positive by applying a minor pose perturbation:
\begin{flalign} \small
\begin{aligned}
\label{eq5}
S_j^+ = S_j \oplus u, \quad u \sim \mathcal{U}(-B,B),
\end{aligned}
\end{flalign}
where $B$ defines the perturbation boundary for translation and orientation, and $\oplus$ denotes pose composition in $SE(2)$. The local floorplan structure cropped at $S_j^+$ is encoded as $\mathbf{g}_j^+$. This slack positive establishes a pose neighborhood around the ground truth. It encourages adjacent poses to maintain high visual-geometric compatibility, thereby promoting local pose smoothness and robustness to minor pose variations.

\textbf{Position-Level Negatives for Spatial Separability.}
We construct position-level negatives $(\mathbf{f}_j,\mathbf{g}_m^{p-})$ by sampling candidate poses with translations distant from the ground-truth location while maintaining comparable orientations. These negatives often capture visually similar local structures in different physical locations. Penalizing these samples suppresses high compatibility scores for spatially distant candidates and directly enhances spatial separability.

\textbf{Orientation-Level Negatives for Angular Discriminability.}
We construct orientation-level negatives $(\mathbf{f}_j,\mathbf{g}_m^{a-})$ by altering the viewing orientation at the correct physical location. These samples force the model to distinguish candidates with accurate translations but incorrect headings. This constraint explicitly improves angular discriminability and mitigates directional ambiguity.

\textbf{Cross-Floorplan Negatives.}
We retain standard cross-floorplan negatives by treating floorplan features from other mini-batch samples as generic negatives. They provide coarse cross-scene separability, while the proposed position-level and orientation-level negatives explicitly target structural aliasing within the same floorplan.

We optimize the visual-geometric compatibility through a single-positive \textbf{P}ose-\textbf{S}tructured \textbf{C}ontrastive (PSC) loss. For concise notation, we define the positive score and three negative partition terms as follows:
\begin{flalign} \small
\begin{aligned}
\label{eq6}
z_j^+ &= \exp(\operatorname{sim}(\mathbf{f}_j,\mathbf{g}_j^+)/\tau),
Z_j^c &= \sum_{m \in \mathcal{N}_j^c}
\exp(\operatorname{sim}(\mathbf{f}_j,\mathbf{g}_m^{c-})/\tau), \\
Z_j^p &= \sum_{m \in \mathcal{N}_j^p}
\exp(\operatorname{sim}(\mathbf{f}_j,\mathbf{g}_m^{p-})/\tau), 
Z_j^a &= \sum_{m \in \mathcal{N}_j^a}
\exp(\operatorname{sim}(\mathbf{f}_j,\mathbf{g}_m^{a-})/\tau).
\end{aligned}
\end{flalign}
Here, $\mathcal{B}$ denotes the training batch, $\mathbf{g}_j^+$ is the slack positive, and $\mathcal{N}_j^c$, $\mathcal{N}_j^p$, and $\mathcal{N}_j^a$ denote the cross-floorplan, position-level, and orientation-level negatives for the $j$-th query, respectively. As shown in Fig. \ref{fig2}(b), visual features $\mathbf{f}$ and floorplan features $\mathbf{g}$ are encoded by a frozen DINOv2-S model \cite{oquab2023dinov2} and a trainable ResNet-18 \cite{he2016deep}, respectively. $\tau$ represents the temperature coefficient. The final contrastive objective is then written as:
\begin{flalign} \small
\begin{aligned}
\label{eq7}
\mathcal{L}_{PSC}
= - \frac{1}{|\mathcal{B}|}
\sum_{j \in \mathcal{B}}
\log
\frac{z_j^+}{z_j^+ + Z_j^c + Z_j^p + Z_j^a}.
\end{aligned}
\end{flalign}

\textbf{$SE(2)$-Aware Candidate Refinement and DisCo Reranking.}
During inference (Fig. \ref{fig2}(c)), we select the Top-$K$ candidate poses with the highest likelihoods from the DAFPM predicted by the RRP. We treat the DAFPM as a high-recall geometric proposal distribution rather than a definitive prediction. Because high-likelihood candidates frequently cluster around a few local modes, directly reranking all Top-$K$ candidates introduces severe computational redundancy.

To extract representative FLoc hypotheses, we perform a greedy $SE(2)$-aware candidate refinement. We sort all candidates by their DAFPM likelihoods. The algorithm iteratively retains the highest-scoring unassigned candidate and suppresses redundant neighbors within its local $SE(2)$ basin:
\begin{flalign} \small
\begin{aligned}
\label{eq8}
d_{SE(2)}(S_i,S_j)
=
\sqrt{
\frac{\|\mathbf{t}_i-\mathbf{t}_j\|_2^2}{\sigma_t^2}
+
\lambda_\theta
\frac{\Delta(\theta_i,\theta_j)^2}{\sigma_\theta^2}
},
\end{aligned}
\end{flalign}
where $\mathbf{t}_i=(x_i,y_i)$ and $\Delta(\theta_i,\theta_j)=\operatorname{wrap}(\theta_i-\theta_j)$. A candidate $S_j$ is suppressed if $d_{SE(2)}(S_i,S_j)\le \rho$. The retained set $\mathcal{R}$ subsequently consists only of spatially and angularly distinct representative candidates.

For each retained candidate $R_m\in\mathcal{R}$, we crop the corresponding local floorplan structure and process it through the learned ResNet-18 encoder. The visual-geometric compatibility is computed as:
\begin{flalign} \small
\begin{aligned}
\label{eq9}
c_m=\operatorname{sim}(\mathbf{f}_{\mathcal{I}},\mathbf{g}_m).
\end{aligned}
\end{flalign}
We then fuse the geometry-based likelihood with the learned compatibility via log-linear reranking:
\begin{flalign} \small
\begin{aligned}
\label{eq10}
P_f(R_m|\mathcal{I},F)
=
\operatorname{softmax}_{R_m\in\mathcal{R}}
\left(
\log P_d(R_m|\mathcal{I},F)+\alpha c_m
\right).
\end{aligned}
\end{flalign}
The final pose is selected as
\begin{flalign} \small
\begin{aligned}
\label{eq11}
\hat{S}_{\mathcal{I},F}
=
\arg\max_{R_m\in\mathcal{R}}
P_f(R_m|\mathcal{I},F).
\end{aligned}
\end{flalign}
This refinement process forces DisCo-FLoc to rerank distinct ambiguity modes rather than redundant neighbors. It significantly reduces computational overhead while preserving parallel feature extraction for efficient real-time inference.

\vspace{-0.3cm}
\section{Experiments}

\subsection{Experimental Setup} \label{4.1}

\textbf{Datasets.} Following existing studies \cite{chen2024f3loc,chen20253dp,chen2025perspective}, we first employ two Gibson \cite{xia2018gibson} datasets, Gibson(g) and Gibson(f), to evaluate our visual FLoc method. We follow the data split in F$^{3}$Loc \cite{chen2024f3loc}, including 108 training scenes, 9 validation scenes, and 9 test scenes. The horizontal \textbf{F}ield \textbf{O}f \textbf{V}iew (FOV) of the images in the Gibson datasets is 108$^\circ$. The images feature upright camera poses and low to medium occlusion. The resolution of the floorplans extracted from the Gibson datasets is 0.1 m. Gibson(g) consists of general motions, including in-place steering motions, and contains 49,558 sequential views. Gibson(f) consists of only forward motions and contains 24,779 sequential views. Therefore, Gibson(g) is intuitively more challenging than Gibson(f).

In addition, we use the challenging Structured3D(full) \cite{zheng2020structured3d} dataset for comparative studies. Structured3D(full) is a photorealistic dataset containing 3,296 fully furnished indoor environments with a total of 78,453 perspective images. The heavy presence of furniture, 3D object occlusions, and repetitive indoor layouts makes this dataset particularly challenging, often leading to severe structural aliasing and multi-modal localization ambiguity under geometry-only matching. We first compare our method with baselines under the F$^{3}$Loc framework, training and evaluating entirely without semantic information. To further evaluate the effectiveness of DisCo-FLoc, we also compare it against strong baselines under the SemRayLoc \cite{grader2025supercharging} framework. These baselines utilize explicit semantic annotations, such as doors, windows, and walls, to provide additional cues for resolving ambiguous localization candidates. Notably, we use monocular rather than panoramic images, with a horizontal FOV of 80$^\circ$. The images feature non-upright camera poses and low to medium degrees of occlusion. The resolution of the floorplans extracted from the Structured3D(full) dataset is 0.02 m. Official data splits are used for all model training and evaluation.

\textbf{Baselines.} We compare our method against three categories of baselines. \textit{The implementation details of our method and all baselines can be found in the appendix.}

\textbf{(1) Early Methods.} PF-net \cite{karkus2018particle} proposes a particle filter specialized for visual FLoc, using an observation model to learn similarities between images and map patches. MCL \cite{dellaert1999monte} serves as a classical framework for 2D localization on pure geometric maps. LASER \cite{min2022laser} represents the floorplan as a point cloud, aggregates features of visible points for each pose, and compares them with query image features. \textbf{(2) Strong Baselines under the F$^{3}$Loc Framework.} F$^{3}$Loc \cite{chen2024f3loc} is a representative probabilistic method that combines a ray-based observation module with histogram filtering. 3DP \cite{chen20253dp} injects 3D geometric priors into the F$^{3}$Loc framework to improve localization accuracy without semantic labels. RSK \cite{chen2025perspective} employs unsupervised room-style knowledge learning to mitigate FLoc ambiguity. 3DP \& RSK adaptively leverages both geometric priors and room-style knowledge (see supplementary material for details). While effective, these methods are not specifically designed to distinguish structurally similar but spatially distant or orientation-inconsistent candidates through pose-structured contrastive supervision. \textbf{(3) Strong Baselines under the SemRayLoc Framework.} SemRayLoc$_s$ \cite{grader2025supercharging} leverages sparse semantic priors in the floorplan to predict semantic rays, using structural-semantic probability volumes to reduce geometric ambiguity. By adapting visual pre-trainings from 3DP and RSK to SemRayLoc$_s$, three enhanced variants are derived: + 3DP, + RSK, and + 3DP \& RSK (see supplementary material). SemRayLoc$_r$ further introduces room-type labels to SemRayLoc$_s$ to provide additional semantic cues for resolving localization ambiguity.

\textbf{Metrics.} Following existing work \cite{chen2024f3loc}, we report recall metrics computed at positional localization accuracies of 0.1 m, 0.5 m, and 1 m. To evaluate both positional and angular localization accuracy, we also report recall under a joint threshold of 1 m positional error and 30$^\circ$ orientation error. Recall is calculated as the percentage of predictions falling within these thresholds. 

\vspace{-0.3cm}
\subsection{Comparisons with SOTA Methods} \label{4.2}

We first evaluate DisCo-FLoc against existing approaches on the Gibson datasets (Tab. \ref{table1}). Even without contrastive disambiguation (Ours w/o Dis.), our method outperforms strong baselines such as 3DP and RSK. This validates that the depth-aware RRP effectively suppresses 3D visual clutter and generates reliable geometry-aware pose candidates. By incorporating the pose-structured contrastive disambiguation module, DisCo-FLoc achieves state-of-the-art performance across all metrics.

The highly cluttered Structured3D(full) dataset further highlights the necessity of our disambiguation module (Tab. \ref{table2}). Due to repetitive furnishings and complex layouts, the initial geometric matching often results in multi-modal probability distributions. Consequently, Ours w/o Dis. yields modest results. The full DisCo-FLoc framework resolves this by suppressing ambiguous candidates, which significantly improves spatial separability and angular discriminability. Notably, our method substantially narrows the performance gap between purely positional localization (R@1 m: 74.3\%) and joint positional-directional localization (R@1 m 30$^\circ$: 73.6\%), demonstrating robust heading estimation.

Finally, we compare DisCo-FLoc against semantic-based frameworks such as SemRayLoc. While these baselines require explicit semantic annotations to resolve localization ambiguities, DisCo-FLoc relies purely on geometric floorplan information. Despite this, it surpasses the strongest semantic-based baseline (SemRayLoc$_s$ + 3DP \& RSK) across all metrics. This confirms that our spatially-perturbed contrastive learning strategy learns robust visual-geometric compatibility. It enables reliable disambiguation and alleviates structural aliasing without requiring costly semantic labels.

\begin{table*}[!t] \small
\centering
\caption{Comparative studies of DisCo-FLoc and baselines on the Gibson(f) and Gibson(g) datasets.}
\label{table1}
\setlength{\tabcolsep}{2.0mm}{
\begin{tabular}{l c c c c | c c c c }
\hline
\hline
\multirow{2}{*}{\textbf{Method \tiny{(Venue)}}} &
\multicolumn{4}{c}{\textbf{Gibson(f) R@}} & \multicolumn{4}{c}{\textbf{Gibson(g) R@}} \\
\cline{2-9}
& \textbf{0.1 m}$\uparrow$  & \textbf{0.5 m}$\uparrow$  & \textbf{1 m}$\uparrow$ & \textbf{1 m 30$^\circ$}$\uparrow$ & \textbf{0.1 m}$\uparrow$ & \textbf{0.5 m}$\uparrow$ & \textbf{1 m}$\uparrow$  & \textbf{1 m 30$^\circ$}$\uparrow$ \\
\hline
PF-net\tiny{(CoRL 2018)} & 0 & 2.0 & 6.9 & 1.2 & 1.0 & 1.9 & 5.6 & 1.9 \\
MCL\tiny{(ICRA 1999)} & 1.6 & 4.9 & 12.1 & 8.2 & 2.3 & 6.2 & 9.7 & 7.3 \\
LASER\tiny{(CVPR 2022)} & 0.4 & 6.7 & 13.0 & 10.4 & 0.7 & 7.0 & 11.8 & 9.5 \\
F$^{3}$Loc\tiny{(CVPR 2024)} & 4.7 & 28.6 & 36.6 & 35.1 & 4.3 & 26.7 & 33.7 & 32.3 \\
3DP\tiny{(ACM MM 2025)} & 5.3 & 33.2 & 39.8 & 38.4 & 9.4 & 37.4 & 43.1 & 41.5 \\
RSK\tiny{(AAAI 2026)} & 8.3 & 38.5 & 45.3 & 43.6 & 8.7 & 36.4 & 42.3 & 40.4 \\
3DP \& RSK & 10.9 & 42.7 & 47.9 & 46.5 & 10.7 & 38.8 & 44.4 & 42.8 \\
\hline
Ours w/o Dis. & 11.8 & 45.0 & 49.6 & 48.3 & 12.3 & 45.0 & 49.9 & 48.2 \\
Ours (DisCo-FLoc) & \textbf{13.1} & \textbf{50.9} & \textbf{56.7} & \textbf{55.4} & \textbf{12.4} & \textbf{48.0} & \textbf{55.6} & \textbf{54.0} \\

\hline
\hline
\end{tabular}}
\end{table*}

\begin{table}[!t] \small
\centering
\caption{Comparative studies of DisCo-FLoc and baselines on the Structured3D(full) dataset. Oracle indicates FLoc using GT geometric and semantic rays, where semantics include doors, windows, and walls. Sem. indicates whether semantics are used.}
\label{table2}
\setlength{\tabcolsep}{4.0mm}{
\begin{tabular}{l | c c c c | c}
\hline
\hline
\multirow{2}{*}{\textbf{Method \tiny{(Venue)}}} &
\multicolumn{4}{c}{\textbf{Structured3D(full) R@}} & \multirow{2}{*}{\textbf{Sem.}} \\
\cline{2-5}
& \textbf{0.1 m}$\uparrow$  & \textbf{0.5 m}$\uparrow$  & \textbf{1 m}$\uparrow$  & \textbf{1 m 30$^\circ$}$\uparrow$ & \\
\hline
PF-net\tiny{(CoRL 2018)} & 0.2 & 1.3 & 3.2 & 0.9 & \multirow{7}{*}{no}  \\
MCL\tiny{(ICRA 1999)} & 1.3 & 5.2 & 7.8 & 6.4 & \\
LASER\tiny{(CVPR 2022)} & 0.7 & 6.4 & 10.4 & 8.7 &  \\
F$^{3}$Loc\tiny{(CVPR 2024)} & 1.5 & 14.6 & 22.4 & 21.3 &  \\
3DP\tiny{(ACM MM 2025)} & 5.6 & 27.4 & 55.5 & 24.0 &  \\
RSK\tiny{(AAAI 2026)} & 6.4 & 28.6 & 56.9 & 25.2 &  \\
3DP \& RSK & 6.7 &  26.8 & 54.7 & 24.2 & \\
\hline
SemRayLoc$_s$\tiny{(ICCV 2025)} & 5.4 & 41.9 & 53.5 & 52.6 & \multirow{5}{*}{yes} \\
\  + 3DP\tiny{(ACM MM 2025)} & 5.5 & 46.6 & 56.7 & 56.2 &  \\
\  + RSK\tiny{(AAAI 2026)} & 6.2 & 48.1 & 59.9 & 58.8 &  \\
\  + 3DP \& RSK & 7.1 & 48.9 & 61.5 & 60.0 &  \\
SemRayLoc$_r$\tiny{(ICCV 2025)} & 5.7 & 45.5 & 58.8 & 57.5 &  \\
\hline
Ours w/o Dis. & 5.5 & 34.2 & 40.4 & 39.3 & \multirow{2}{*}{no}\\
Ours (DisCo-FLoc) & \textbf{9.5} & \textbf{63.9} & \textbf{74.3} & \textbf{73.6} & \\
\hline
Oracle w/ sem & 61.0 & 93.9 & 94.9 & 94.6 & yes \\
\hline
\hline
\end{tabular}}
\vspace{-0.5cm}
\end{table}

\begin{table}[t]
\vspace{-0.2cm}
\centering
\small
\caption{
Ablation studies on Structured3D(full). 
(a) evaluates inner-floorplan position-level negatives (I-Pos-N) and orientation-level negatives (Ori-N), while retaining cross-floorplan negatives. 
(b) evaluates slack positive sampling with positional perturbation (P-Pert) and angular perturbation (A-Pert), while retaining all negatives.
}
\label{tab:ablation}
\setlength{\tabcolsep}{1.2mm}

\begin{minipage}{0.48\textwidth} \small
\centering
\textbf{(a) Within-floorplan hard negatives}
\vspace{0.08cm}

\begin{tabular}{cc|cccc} 
\hline
\hline
\multicolumn{2}{c|}{\textbf{Ablations}} &
\multicolumn{4}{c}{\textbf{Structured3D(full) R@}} \\
\hline
I-Pos-N & Ori-N 
& \textbf{0.1 m}$\uparrow$ 
& \textbf{0.5 m}$\uparrow$ 
& \textbf{1 m}$\uparrow$ 
& \textbf{1 m 30$^\circ$}$\uparrow$ \\
\hline
\ding{55} & \ding{55} & 9.0 & 60.4 & 70.7 & 70.0 \\
\checkmark & \ding{55} & 9.5 & 62.4 & 73.5 & 72.8 \\
\ding{55} & \checkmark & 9.5 & 62.8 & 73.7 & 73.0 \\
\checkmark & \checkmark & \textbf{9.5} & \textbf{63.9} & \textbf{74.3} & \textbf{73.6} \\
\hline
\hline
\end{tabular}
\vspace{-0.5cm}
\end{minipage}
\hfill
\begin{minipage}{0.48\textwidth}
\centering
\textbf{(b) Slack positives}
\vspace{0.08cm}

\setlength{\tabcolsep}{0.5mm}
\begin{tabular}{cc|ccccc}
\hline
\hline
\multicolumn{2}{c|}{\textbf{Ablations}} &
\multicolumn{4}{c}{\textbf{Structured3D(full) R@}} \\
\hline
P-Pert & A-Pert
& \textbf{0.1 m}$\uparrow$ 
& \textbf{0.5 m}$\uparrow$ 
& \textbf{1 m}$\uparrow$ 
& \textbf{1 m 30$^\circ$}$\uparrow$ \\
\hline
\checkmark & \ding{55} & 8.6 & 54.5 & 63.4 & 62.7 \\
\ding{55} & \checkmark & 8.9 & 57.8 & 67.0 & 66.1 \\
\checkmark & \checkmark & \textbf{9.5} & \textbf{63.9} & \textbf{74.3} & \textbf{73.6} \\
\hline
\hline
\end{tabular}
\vspace{-0.5cm}
\end{minipage}

\vspace{-0.2cm}
\end{table}

\vspace{-0.3cm}
\subsection{Ablation and Parametric Studies} \label{5.3}

We conduct comprehensive ablation and parametric studies on the challenging Structured3D(full) dataset to analyze the contribution of each algorithmic component.

\begin{wrapfigure}{r}{0.55\linewidth}
\vspace{-0.2cm}
\centering
\includegraphics[width=1.0\linewidth]{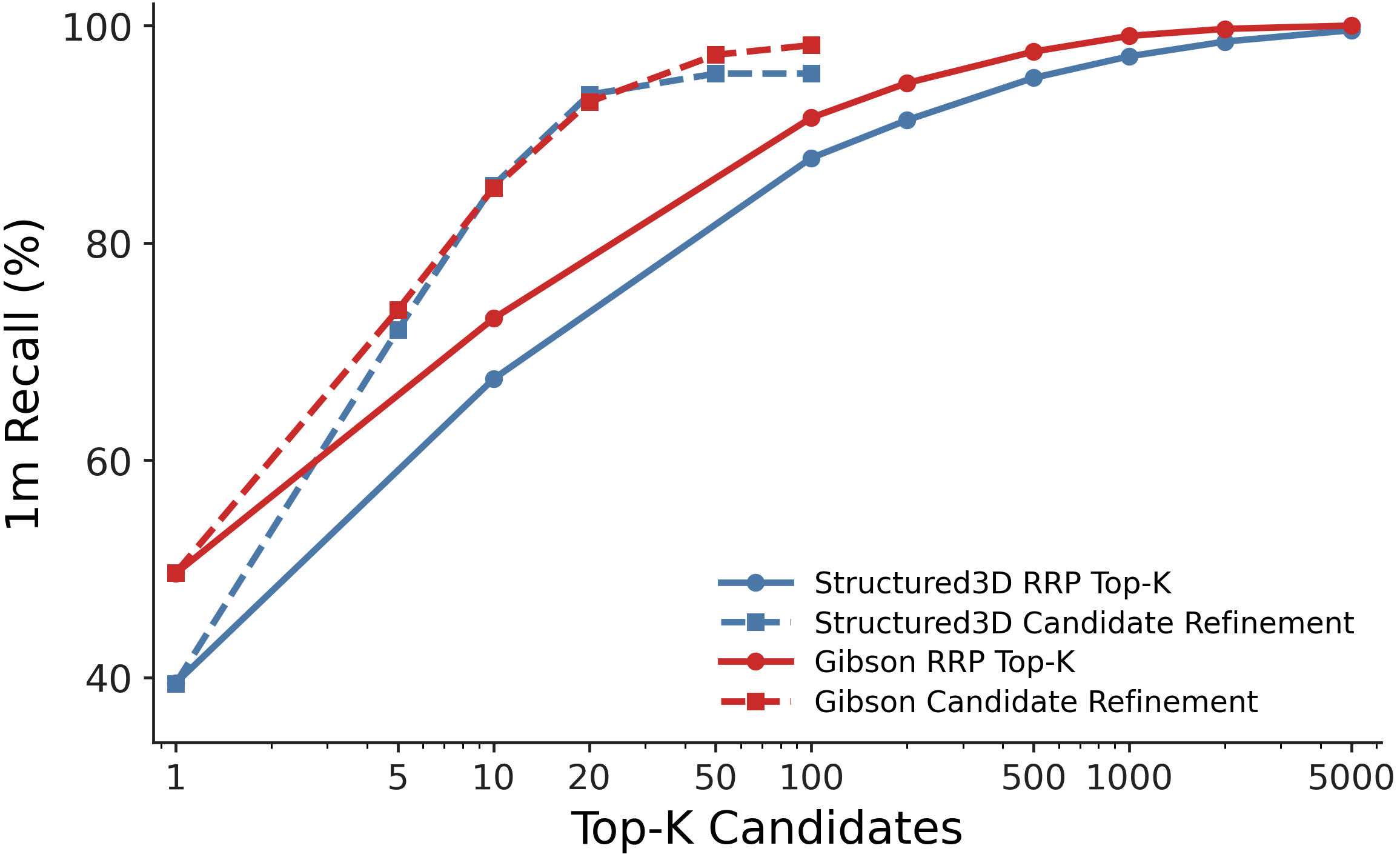}
\vspace{-0.5cm}
\caption{
Candidate coverage before and after $SE(2)$-aware refinement. Solid and dashed curves denote Top-$K$ DAFPM candidates and refined representative hypotheses, respectively.
}
\vspace{-0.4cm}
\label{fig:topk_coverage}
\end{wrapfigure}

\textbf{Candidate Coverage and Representative Hypothesis Extraction.}
We first evaluate the candidate quality of the depth-aware RRP. As shown in Fig. \ref{fig:topk_coverage}, the Top-1000 DAFPM candidates achieve R@1 m scores of 99.05\% and 97.17\% on Gibson(f) and Structured3D(full), respectively, confirming that dense-to-ray projection provides a high-recall geometry-based proposal distribution. Meanwhile, the refined representative hypotheses preserve comparable candidate coverage with substantially fewer hypotheses. To enable efficient visual-geometric disambiguation, we extract representative pose hypotheses from distinct high-likelihood modes using $SE(2)$-aware candidate refinement, rather than reranking all Top-1000 candidates. This operation suppresses neighboring candidates within an $SE(2)$ radius, reducing the candidate set to only $\sim$42 representative hypotheses on average while preserving the major ambiguity modes. \textit{We report the inference time of our method in Tab. \ref{tab:runtime} of the appendix.} Our semantic-free approach is not only more computationally efficient than the semantic-based SemRayLoc, but also achieves higher localization accuracy.


\textbf{Dual-Level Negative Constraints.} We then evaluate our contrastive objective for FLoc disambiguation. Tab. \ref{tab:ablation}(a) starts from a baseline with only standard cross-floorplan negatives and progressively adds the two proposed within-floorplan hard negatives. Introducing inner-floorplan position-level negatives (I-Pos-N) improves spatial separability by suppressing physically distant but structurally similar candidates. Similarly, orientation-level negatives (Ori-N) enhance angular discriminability by penalizing incorrect headings at the true location. The full model, which combines both within-floorplan negative types with cross-floorplan negatives, achieves the highest accuracy.

\textbf{Slack Positives.} Tab. \ref{tab:ablation}(b) reports the ablation on slack positives while retaining the full negative set. Incorporating positional perturbation (P-Pert) and angular perturbation (A-Pert) encourages local pose smoothness. While both perturbations improve robustness to minor pose variations, A-Pert yields particularly large gains on the joint positional-directional metric (R@1 m 30$^\circ$). These results verify that our spatially perturbed sampling effectively unifies local pose smoothness, spatial separability, and angular discriminability.

\textbf{Parametric Studies.} Finally, we investigate key hyperparameters for the disambiguation module. Empirical results indicate that a fusion weight of $\alpha = 0.5$ and a local floorplan crop size of 5 m $\times$ 5 m provide the optimal balance between the initial geometric likelihood and the learned pose-aware compatibility. \textit{Detailed experimental results are provided in Tab. \ref{tab:supp_weight} and Tab. \ref{tab:supp_crop} in the appendix.}

\vspace{-0.3cm}
\section{Conclusion and Limitations}
This paper presents DisCo-FLoc, a semantic-free framework designed to mitigate structural aliasing in visual FLoc. Instead of relying on costly semantic annotations, our method integrates depth-aware candidate generation with $SE(2)$-aware contrastive disambiguation. First, we introduce a depth-aware RRP that functions as a dense-to-ray geometric projector. By suppressing vertical visual clutter and transforming monocular RGB observations into 2D ray primitives, the RRP bridges the visual-floorplan modal gap to generate reliable pose candidates. Second, to resolve residual ambiguities, we propose a spatially perturbed contrastive disambiguation objective. By incorporating slack positives alongside dual-level negative constraints, this objective aligns RGB images with local floorplan structures. Consequently, it enhances local pose smoothness, spatial separability, and angular discriminability. Extensive evaluations confirm that DisCo-FLoc achieves SOTA performance and surpasses existing semantic-based methods without using any semantic labels. Furthermore, the minimal gap between purely positional and joint positional-directional accuracy highlights its robust heading estimation capabilities.

\textbf{Limitations.} While DisCo-FLoc successfully resolves structural aliasing without semantic labels, the framework currently operates in two sequential stages. Although the $SE(2)$-aware candidate refinement significantly reduces redundant geometric matching, the inherent dependency between initial candidate generation and contrastive reranking introduces additional inference latency. Future work will explore an end-to-end differentiable architecture. We aim to unify the dense-to-ray projection, candidate refinement, and pose-structured contrastive disambiguation into a single forward pass to further optimize computational efficiency.


\bibliographystyle{plain}
\bibliography{neurips_2026}

\clearpage
\newpage
\appendix

\section{More Experimental Details}

\subsection{Implementation Details of Our Method}
To optimize the dense-to-ray geometric projection (RRP), we employ the Adam optimizer \cite{kingma2014adam} with a constant learning rate of $10^{-4}$ and a batch size of 64. The depth-aware visual encoder (DINOv2-S from Depth Anything V2 \cite{yang2024depth}) is frozen during training to preserve its depth estimation expertise. The remaining components are trained for 50 epochs on an NVIDIA RTX 3090 GPU. Our RRP matches 40 predicted rays to the floorplan to generate the initial geometry-aware pose likelihood map. For ray-depth discretization and candidate ray matching, we follow the same settings as F$^3$Loc \cite{chen2024f3loc}, including the values of $D$, $d_{\text{min}}$, $d_{\text{max}}$, $\gamma$, and the $SE(2)$ pose discretization. 

For the spatially-perturbed contrastive disambiguation module, the model is optimized using Adam with an initial learning rate of $10^{-3}$, a batch size of 32, a weight decay of $5\times 10^{-5}$, and a cosine learning-rate schedule. The model is trained on an NVIDIA RTX 3090 GPU for 20 epochs, identifying the optimal checkpoint via early stopping based on the minimum validation loss. Unless specified otherwise, the crop size for local floorplan structures is 5 m $\times$ 5 m. Cross-floorplan negatives are taken from other mini-batch samples and retained in all ablations. To encourage \textit{pose smoothness} via slack positive sampling, we apply positional and orientational perturbations of 0.5 m and $\pm 0.26$ radians to the GT pose, respectively. To improve \textit{spatial separability}, inner-floorplan position-level negative samples are randomly drawn within a distance range of 1.5 m to 3.0 m from the GT pose. To improve \textit{angular discriminability}, orientation-level negative samples are generated by applying a $180^\circ$ rotation to the GT orientation. During inference, we first extract the top $K = 1000$ candidate poses from the DAFPM to ensure high candidate coverage. We then apply greedy $SE(2)$-aware candidate refinement with $\sigma_t = 0.6$ m, $\sigma_\theta = 30^\circ$, $\lambda_\theta = 1.0$, and $\rho = 1.0$ before DisCo reranking. We set the log-linear reranking weight in Eq.~\ref{eq10} to $\alpha=0.5$.

\subsection{Implementation of 3DP \& RSK}
\label{sec:impl_3dp_rsk}

\begin{figure*}[t]
 \centering
 \includegraphics[width=0.9\linewidth]{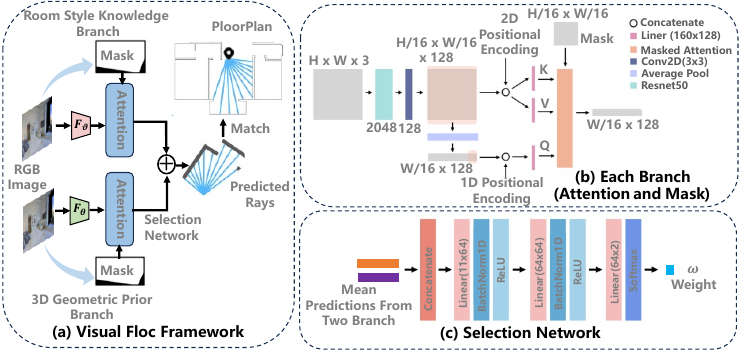}
 \caption{Illustrations of 3DP \& RSK. (a) The pre-trained visual encoders, $F_\theta$ in 3DP and $F_\vartheta$ in RSK, are transferred to the visual FLoc framework for fine-tuning to further fit this task. The visual FLoc framework is a dual-branch model consisting of a 3D geometric prior branch and a RSK branch. (b) and (c) detail the masked attention mechanism and the selection network, respectively.}
 \label{fig:supp_3dp_rsk}
\end{figure*}

3DP \& RSK is achieved by integrating the visual encoders pre-trained in 3DP \cite{chen20253dp} and RSK \cite{chen2025perspective} into a unified FLoc framework.
As shown in Fig.~\ref{fig:supp_3dp_rsk}(a), the fully pre-trained encoders, $F_\theta$ in 3DP and $F_\vartheta$ in RSK, are transferred to the visual FLoc framework for fine-tuning to fit the task, which localizes by finding the pose in the floorplan that has the most similar 2D rays as the prediction.
Our visual FLoc framework is implemented as a dual-branch model consisting of a 3D geometric prior branch and a RSK branch.
In each branch, the image is first aligned with the gravity direction, as done in \cite{chen2024f3loc}.
Then, $F_{\theta}$/$F_{\vartheta}$ and an attention \cite{vaswani2017attention}-based network are used to learn the probability distribution of planar depth over a range of depth hypotheses.
Pixels that become unobservable due to gravity alignment are masked in the attention, as shown in Fig.~\ref{fig:supp_3dp_rsk}(b).
To adaptively leverage 3D geometric priors and RSK based on the current view, a selection network, shown in Fig.~\ref{fig:supp_3dp_rsk}(c), implemented as a multilayer perceptron is adopted to learn a weight $0 \le \omega \le 1$ from the two predictions for adaptive selection:
\begin{equation}
\label{eq:supp_fusion}
\mathbf{P}_{\mathrm{Fusion}} = \omega \mathbf{P}_{\mathrm{3DP}} + (1 - \omega)\mathbf{P}_{\mathrm{RSK}}.
\end{equation}
$\mathbf{P}_{\mathrm{3DP}}$ and $\mathbf{P}_{\mathrm{RSK}}$ denote the probability distributions of planar depth from the 3D geometric prior branch and the RSK branch, respectively.
The expectation of $\mathbf{P}_{\mathrm{Fusion}}$ provides the final prediction of 2D rays.
$\omega$ is manually specified as 1 and 0, implying that only 3D geometric priors and RSK are used, respectively.
For the training of FLoc models, we optimize an L1 loss and a cosine-similarity-based shape loss:
\begin{equation}
\label{eq:supp_floc_loss}
\mathcal{L}_{\mathrm{FLoc}} =
\left\lVert \mathbf{d} - \mathbf{d}^{*} \right\rVert_1
+ 1 - \frac{\mathbf{d}^{\top}\mathbf{d}^{*}}
{\max\left\{\lVert \mathbf{d}\rVert_2 \lVert \mathbf{d}^{*}\rVert_2,\epsilon \right\}},
\end{equation}
where $\mathbf{d}$ and $\mathbf{d}^{*}$ are predicted and GT 2D-ray depths, respectively.
$\epsilon$ is a small constant to prevent division by zero.

\subsection{Variants of SemRayLoc}
\label{sec:variants_semrayloc}

\begin{figure*}[t]
 \centering
 \includegraphics[width=1.0\linewidth]{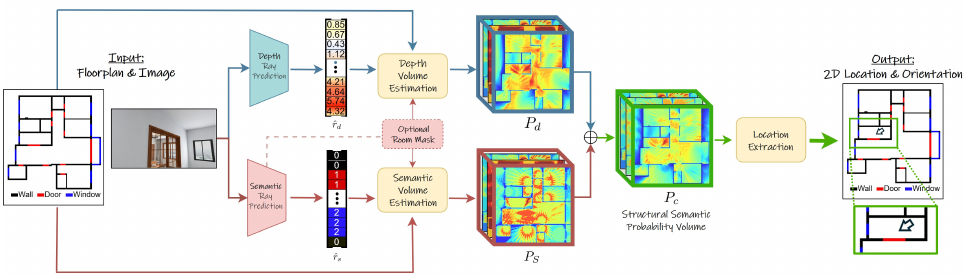}
 \caption{The pipeline of SemRayLoc from \cite{grader2025supercharging}.}
 \label{fig:supp_semrayloc}
\end{figure*}

Similar to the implementation of 3DP \& RSK in Subsection~\ref{sec:impl_3dp_rsk}, \textbf{SemRayLoc$_s$ + 3DP} is implemented by replacing the depth and semantic ray encoders in Fig.~\ref{fig:supp_semrayloc} with the fully pre-trained visual encoder $F_\theta$ from 3DP.
Similarly, \textbf{SemRayLoc$_s$ + RSK} is implemented by replacing the depth and semantic ray encoders in Fig.~\ref{fig:supp_semrayloc} with the fully pre-trained visual encoder $F_\vartheta$ from RSK.
\textbf{SemRayLoc$_s$ + 3DP \& RSK} is implemented by replacing the depth and semantic ray encoders with $F_\theta$ and $F_\vartheta$, respectively.

\section{Additional Evaluation for Long-Sequence Trajectory Tracking}

In this section, we compare our method with existing SOTA methods on the long-sequence trajectory tracking task using the Gibson(t) dataset collected by F$^3$Loc \cite{chen2024f3loc}.
Gibson(t) consists of 118 long image sequences, each containing 280 $\sim$ 5152 image frames.
The \textbf{R}oot-\textbf{M}ean-\textbf{S}quare \textbf{E}rror (RMSE) over the last 10 frames is employed to measure the accuracy of sequential trajectory tracking when localization is successful (RMSE(S)) and in all cases (RMSE(A)).
Technically, we combine the histogram filter proposed by F$^3$Loc with our method and perform visual FLoc with 100 historical frames.

As shown in Tab.~\ref{tab:supp_tracking}, our method improves recall by 2.7\% compared to 3DP at the localization accuracy of 0.2 m.
The reduction in the RMSE(S) metric reflects the robustness of our method in sequential trajectory tracking.
Additionally, our method achieves competitive results in terms of 1 m localization accuracy and RMSE(A).
It is worth noting that F$^3$Loc fusion performs visual FLoc by adaptively utilizing single-frame and multi-frame images, yet it is only competitive on the RMSE(S) metric.
Oracle achieved 100\% Recall at 1 m accuracy by combining GT rays with a histogram filter, demonstrating the potential of visual FLoc.

\begin{table}[t]
\centering
\caption{Comparative studies of long-sequence trajectory tracking methods on the Gibson(t) dataset.}
\label{tab:supp_tracking}
\setlength{\tabcolsep}{2.0mm}
\begin{tabular}{lcccc}
\hline
\hline
\multirow{2}{*}{\textbf{Method \tiny{(Venue)}}} &
\multicolumn{4}{c}{\textbf{Gibson(t) R@}} \\
\cmidrule(lr){2-5}
& \textbf{0.2 m}$\uparrow$ & \textbf{1 m}$\uparrow$ & \textbf{RMSE(S)}$\downarrow$ & \textbf{RMSE(A)}$\downarrow$ \\
\midrule
LASER\tiny{(CVPR 2022)} & - & 59.5 & 0.39 & 1.96 \\
F$^{3}$Loc\tiny{(CVPR 2024)} & 35.1 & 89.2 & 0.18 & 0.88 \\
F$^{3}$Loc fusion\tiny{(CVPR 2024)} & 62.2 & 94.6 & 0.12 & 0.51 \\
3DP\tiny{(ACM MM 2025)} & 70.3 & 97.3 & 0.12 & 0.34 \\
RSK\tiny{(AAAI 2026)} & 59.5 & 94.6 & 0.13 & 0.51 \\
3DP \& RSK & 64.9 & 94.6 & 0.12 & 0.51 \\
\midrule
Ours (DisCo-FLoc) & 73.0 & 94.6 & 0.11 & 0.49 \\
\midrule
Oracle & - & 100.0 & 0.07 & 0.07 \\
\hline
\hline
\end{tabular}
\end{table}

\begin{table}[t]
\centering
\small
\caption{Average single-query inference time (ms).}
\label{tab:runtime}
\setlength{\tabcolsep}{6mm}
\begin{tabular}{l c}
\hline
\hline
\textbf{Method} & \textbf{Time (ms)} $\downarrow$ \\
\hline
F$^3$Loc & 6.4 \\
SemRayLoc & 169.4 \\
RRP & 11.56 \\
RRP + DisCo & 42.34 \\
\hline
\hline
\end{tabular}
\vspace{-0.2cm}
\end{table}

\begin{table}[t]
\centering
\caption{Additional parameter study on the disambiguation weight $\alpha$ on the Structured3D(full) dataset.}
\label{tab:supp_weight}
\setlength{\tabcolsep}{2.0mm}
\begin{tabular}{c|cccc}
\hline
\hline
\textbf{Disam. Weight} & \textbf{0.1 m}$\uparrow$ & \textbf{0.5 m}$\uparrow$ & \textbf{1 m}$\uparrow$ & \textbf{1 m 30$^\circ$}$\uparrow$ \\
\midrule
0.0 & 5.1 & 32.8 & 39.0 & 38.9 \\
0.1 & 9.3 & 62.2 & 72.4 & 71.2 \\
0.3 & 9.7 & 64.6 & 74.0 & 73.5 \\
0.5 & 9.5 & 63.9 & 74.3 & 73.6 \\
0.7 & 9.4 & 63.0 & 73.3 & 63.0 \\
0.9 & 9.2 & 52.0 & 72.9 & 62.5 \\
\hline
\hline
\end{tabular}
\end{table}
As shown in Tab.~\ref{tab:supp_weight}, increasing $\alpha$ from 0.0 substantially improves localization accuracy, confirming the importance of the learned visual-geometric compatibility. The best overall performance is achieved at $\alpha=0.5$, while overly large weights degrade the joint positional-directional metric, suggesting that excessive reliance on the learned compatibility may weaken the stabilizing effect of the geometry-based likelihood.

\begin{table}[t]
\centering
\caption{Additional parameter study on the crop size of floorplan structures on the Structured3D(full) dataset.}
\label{tab:supp_crop}
\setlength{\tabcolsep}{2.0mm}
\begin{tabular}{c|cccc}
\hline
\hline
\textbf{Crop Size} & \textbf{0.1 m}$\uparrow$ & \textbf{0.5 m}$\uparrow$ & \textbf{1 m}$\uparrow$ & \textbf{1 m 30$^\circ$}$\uparrow$ \\
\midrule
3 m $\times$ 3 m & 8.7 & 55.6 & 65.7 & 64.5 \\
5 m $\times$ 5 m & 9.5 & 63.9 & 74.3 & 73.6 \\
7 m $\times$ 7 m & 9.7 & 62.5 & 72.5 & 71.8 \\
\hline
\hline
\end{tabular}
\end{table}

\section{Analysis of Computational Efficiency}

We evaluate the single-query inference time to demonstrate the practical efficiency of our framework (Tab. \ref{tab:runtime}). While pure geometric methods like F$^3$Loc are lightweight (6.4 ms), they suffer from severe structural aliasing and yield limited accuracy (R@1 m: 22.4\%). In contrast, the full DisCo-FLoc framework requires only 42.34 ms per query, making it approximately four times faster than the semantic-based SemRayLoc framework (169.4 ms). SemRayLoc incurs heavy computational costs for explicit semantic extraction and matching. Most importantly, DisCo-FLoc pairs this high efficiency with significant performance gains. Operating in a semantic-free manner, it achieves an absolute improvement of 12.8\% on the R@1 m metric over the strongest semantic-based baseline (Tab. \ref{table2}). These results confirm that our $SE(2)$-aware contrastive strategy successfully maximizes spatial separability and angular discriminability without the severe inference latency typical of semantic-dependent systems.

\section{Additional Parametric Studies}

In this section, we provide additional parametric studies on the disambiguation weight $\alpha$ and the crop size of local floorplan structures. As discussed in the main paper, we evaluate these parameters on the Structured3D(full) dataset. As shown in Fig. \ref{tab:supp_weight}, setting $\alpha$ to 0.3 or 0.5 helps achieve optimal visual FLoc performance. Tab.~\ref{tab:supp_crop} shows that a 5 m $\times$ 5 m crop achieves the best overall performance. Smaller crops may miss sufficient surrounding layout context, whereas larger crops can introduce irrelevant structures from neighboring regions, making visual-geometric matching less discriminative.

\section{More Visualizations}

\begin{figure*}[t]
    \centering
    \setlength{\tabcolsep}{1pt}
    \renewcommand{\arraystretch}{0.9}
    \newcommand{\imgw}{0.17\linewidth}

    \begin{tabular}{c c c c c c}

        \small RGB Input
        & \includegraphics[width=\imgw]{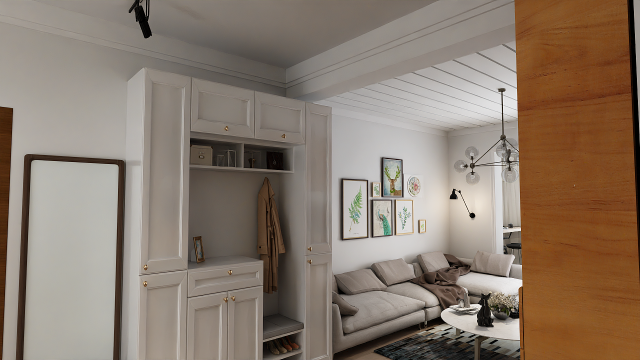}
        & \includegraphics[width=\imgw]{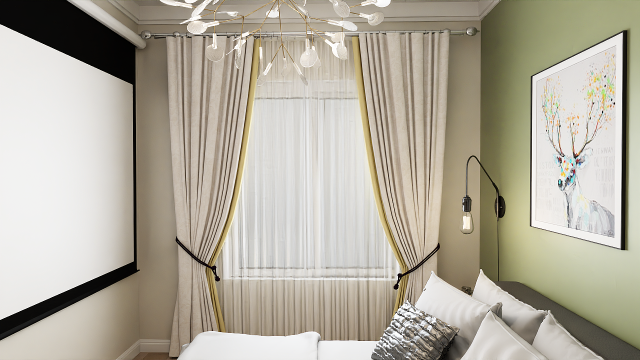}
        & \includegraphics[width=\imgw]{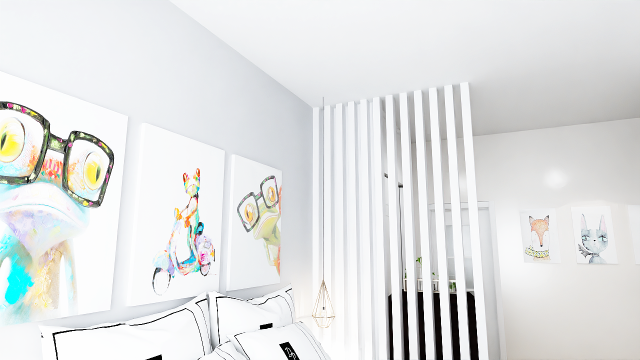}
        & \includegraphics[width=\imgw]{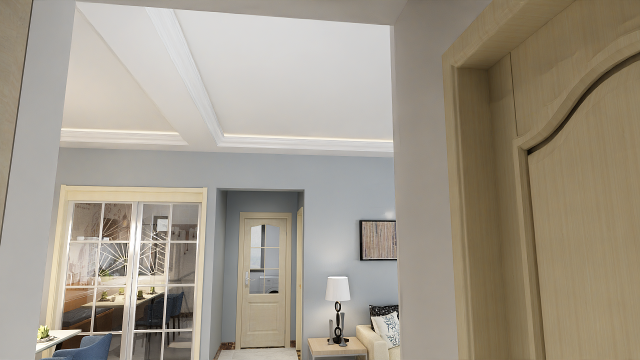}
        & \includegraphics[width=\imgw]{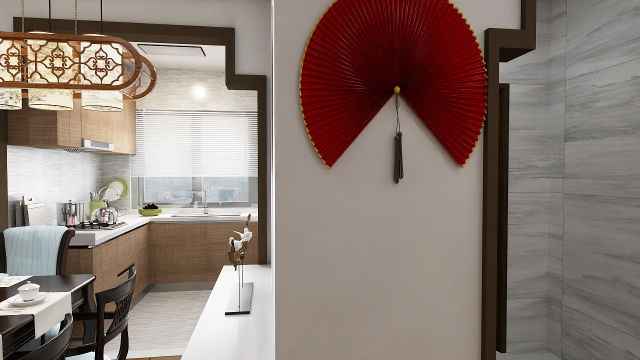} \\

        \small SemRayLoc
        & \includegraphics[width=\imgw]{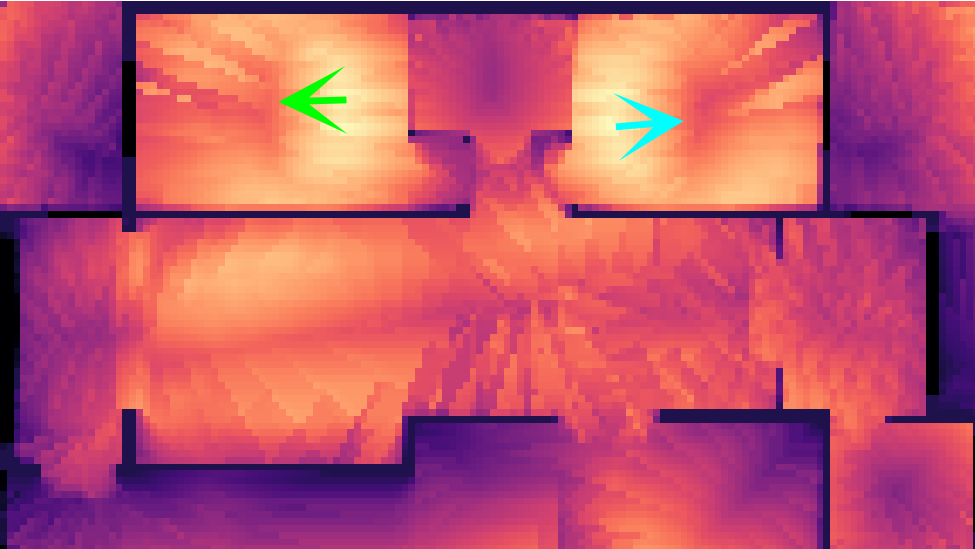}
        & \includegraphics[width=\imgw]{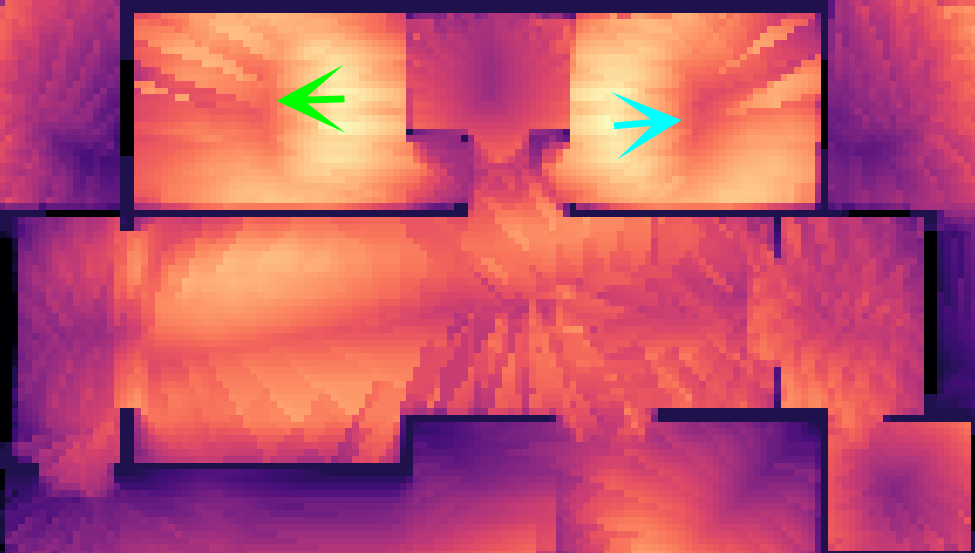}
        & \includegraphics[width=\imgw]{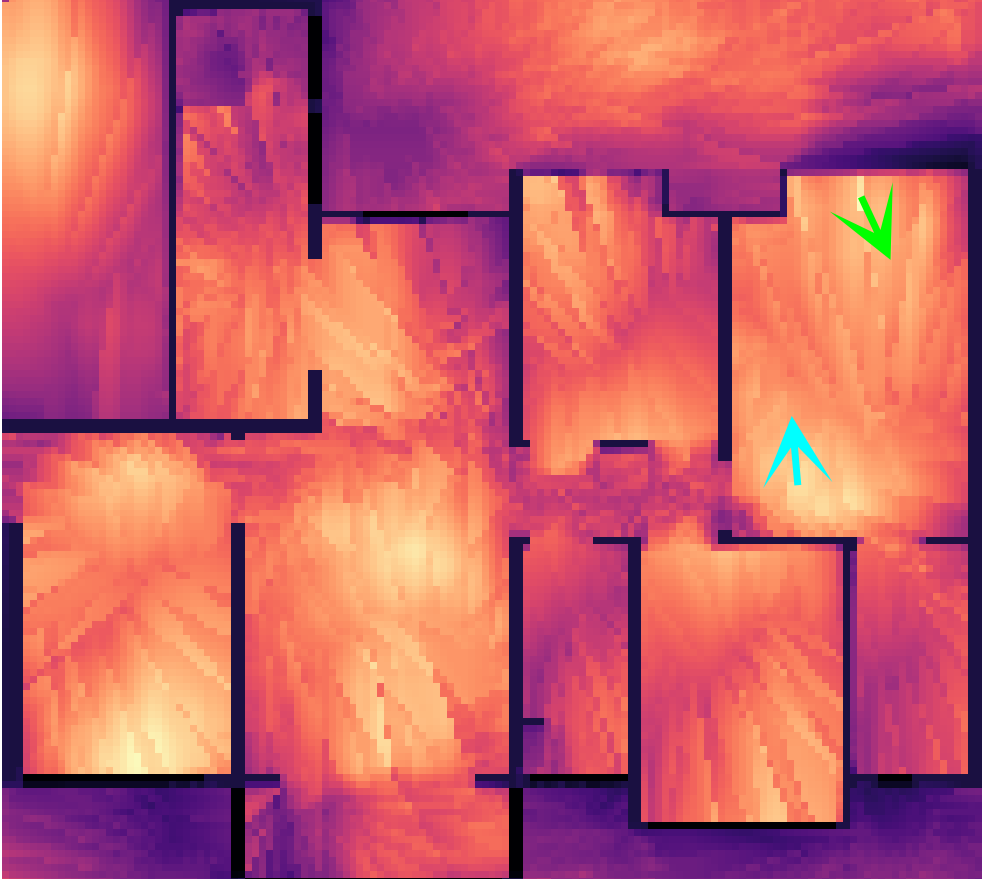}
        & \includegraphics[width=\imgw]{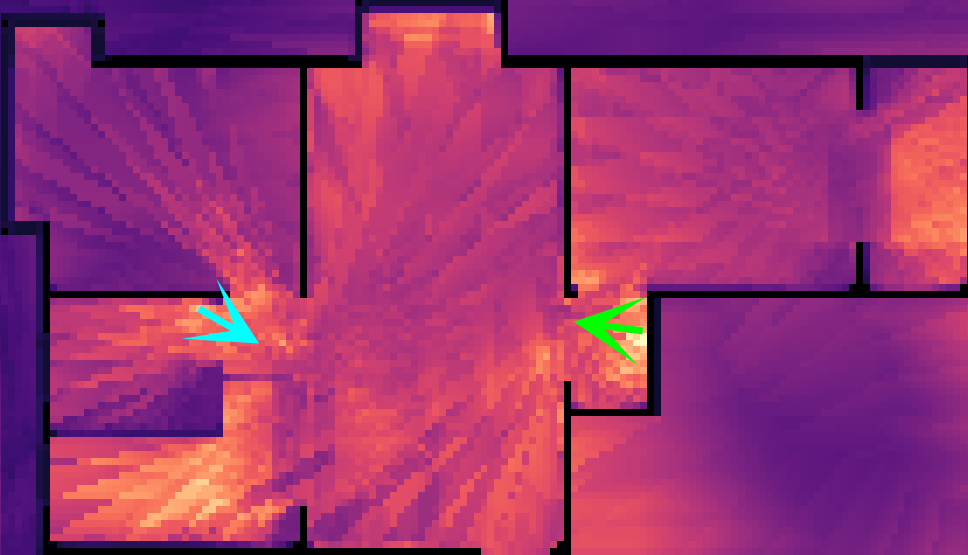}
        & \includegraphics[width=\imgw]{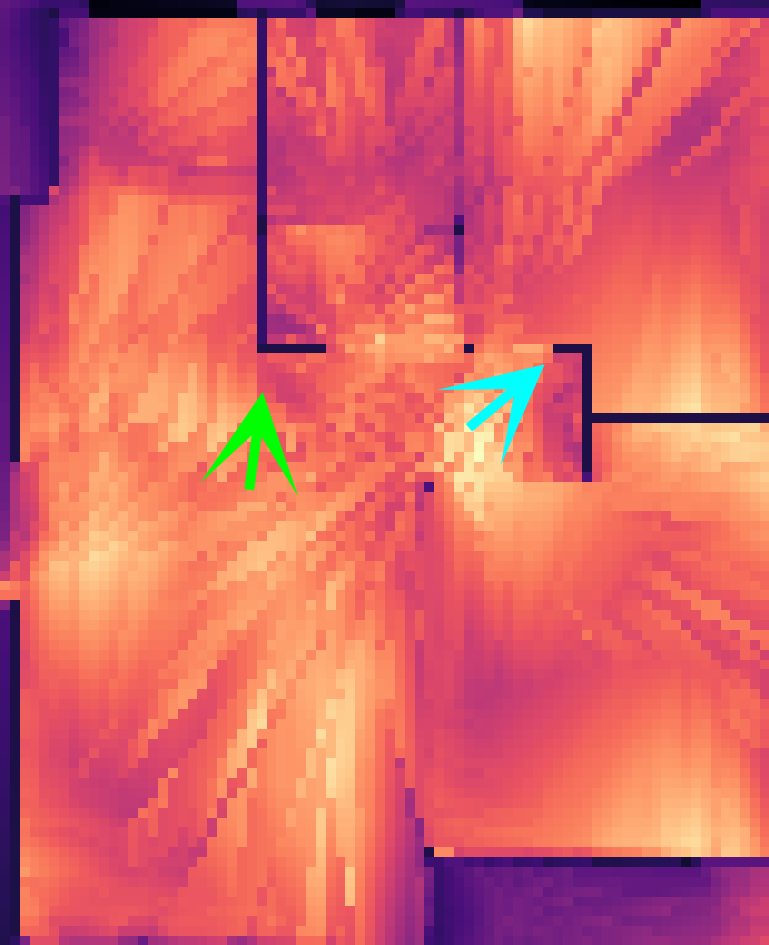} \\

        \small Our RRP
        & \includegraphics[width=\imgw]{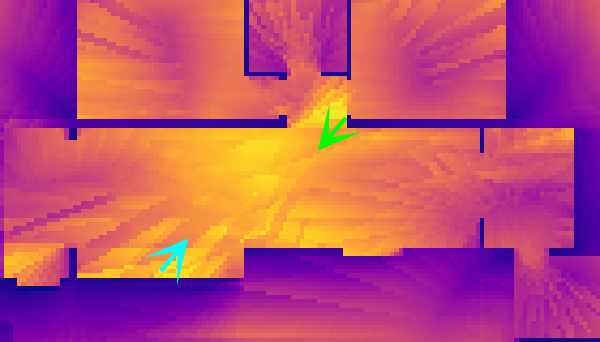}
        & \includegraphics[width=\imgw]{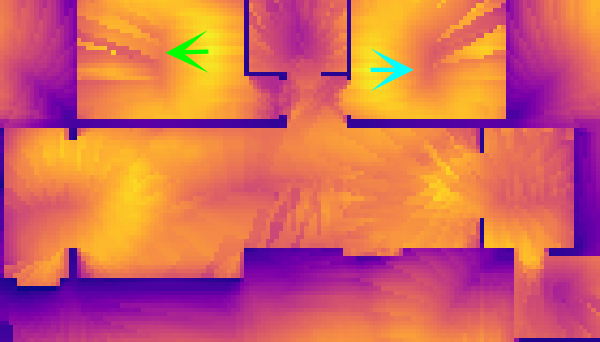}
        & \includegraphics[width=\imgw]{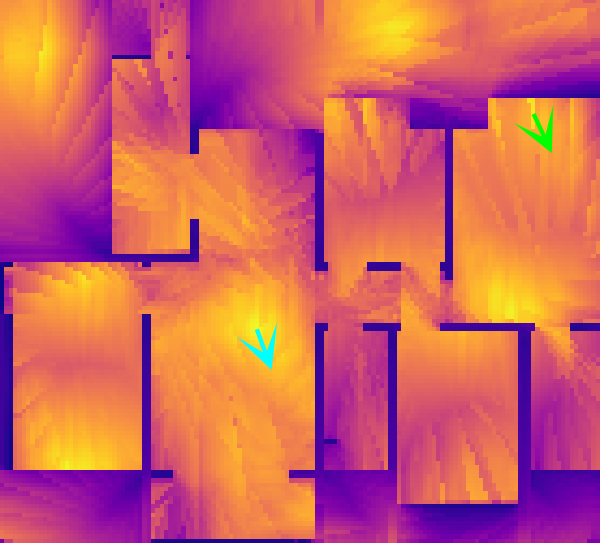}
        & \includegraphics[width=\imgw]{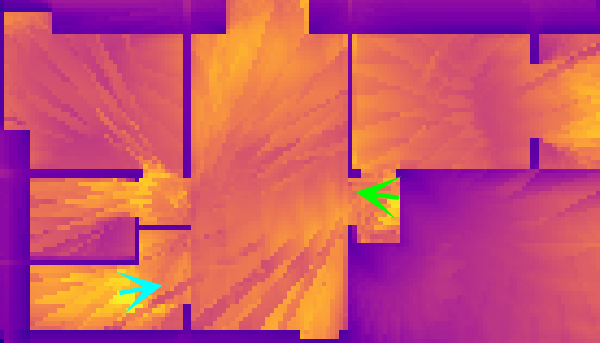}
        & \includegraphics[width=\imgw]{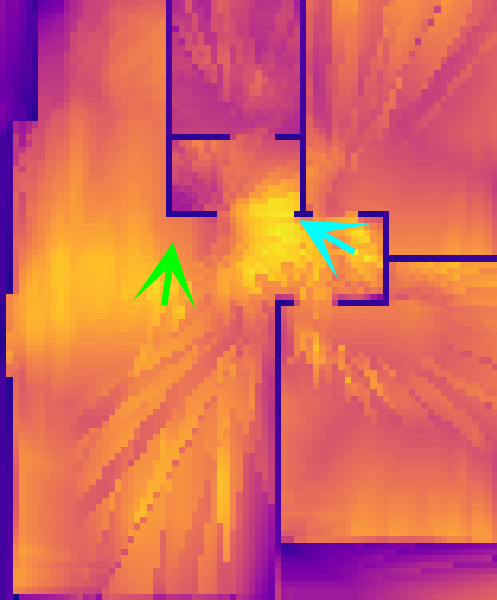} \\

        \small Our DPM
        & \includegraphics[width=\imgw]{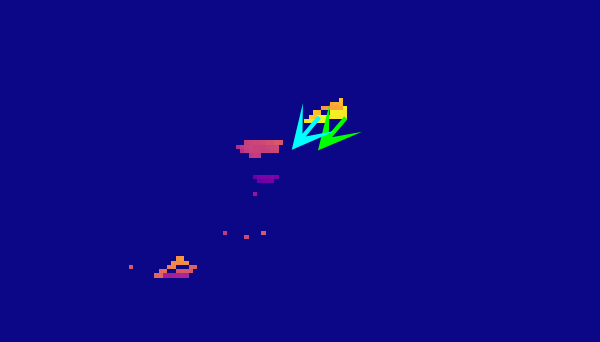}
        & \includegraphics[width=\imgw]{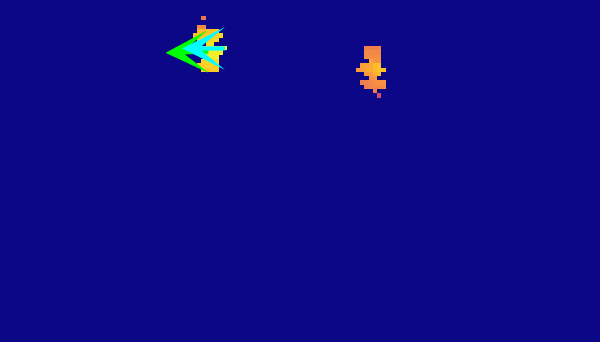}
        & \includegraphics[width=\imgw]{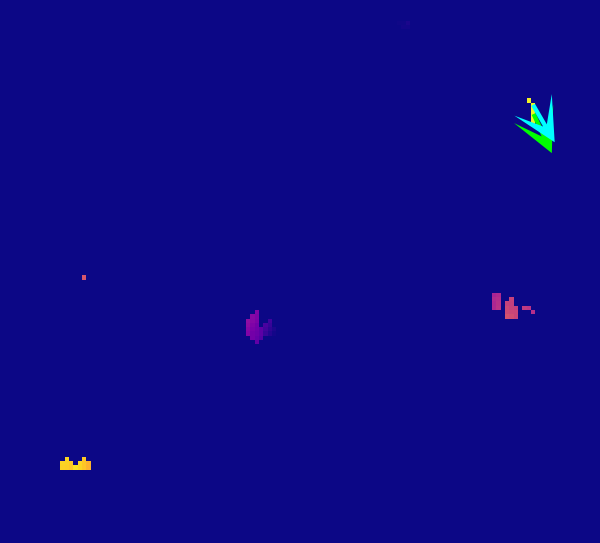}
        & \includegraphics[width=\imgw]{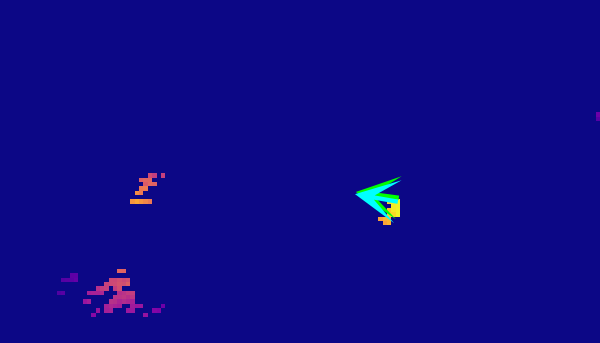}
        & \includegraphics[width=\imgw]{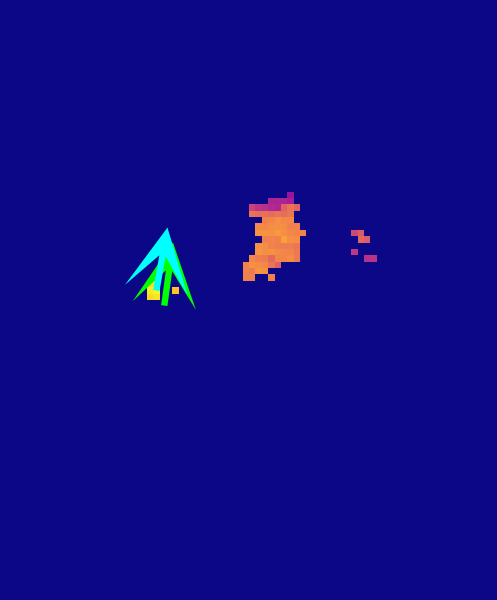} \\

        \small Final FLoc
        & \includegraphics[width=\imgw]{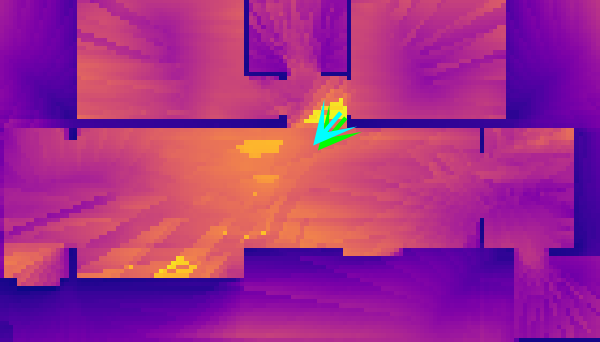}
        & \includegraphics[width=\imgw]{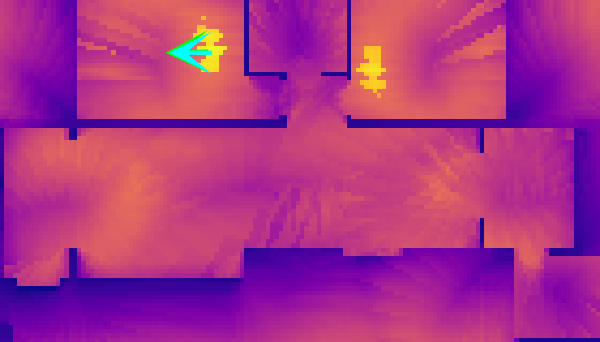}
        & \includegraphics[width=\imgw]{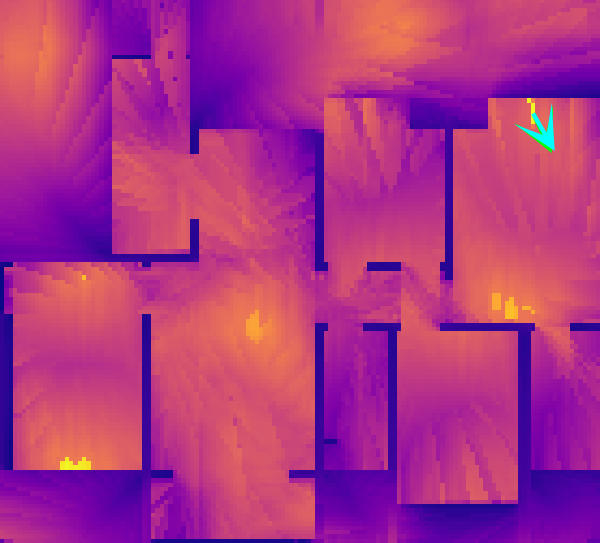}
        & \includegraphics[width=\imgw]{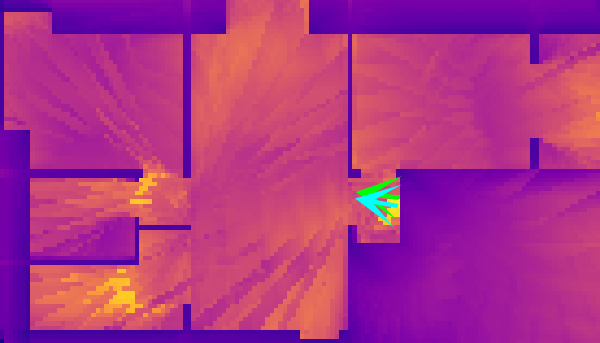}
        & \includegraphics[width=\imgw]{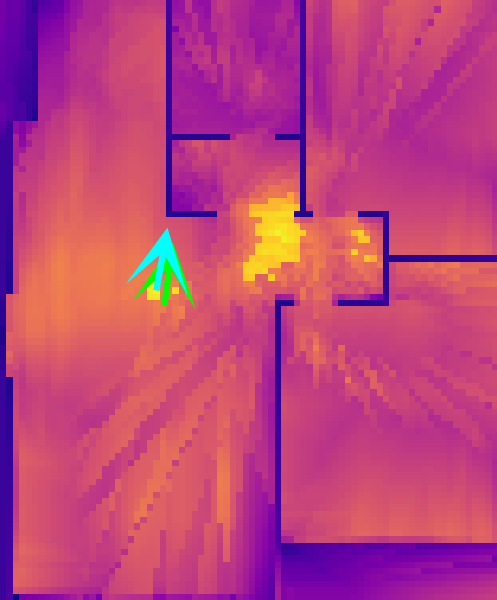} \\
    \end{tabular}

    \caption{Qualitative comparison on 5 scenes.}
    \label{fig:supp_visualization}
\end{figure*}

Fig.~\ref{fig:supp_visualization} shows qualitative comparisons between our method and SemRayLoc \cite{grader2025supercharging}.
Due to the repetitive geometric structures in the floorplan, our RRP also fails when SemRayLoc encounters errors.
However, our visual-geometric contrastive disambiguation effectively addresses localization ambiguities and yields highly accurate final FLoc results.

\paragraph{Broader Impacts.}
This work studies semantic-free visual floorplan localization, which can benefit indoor robotics, augmented reality, assistive navigation, and long-term spatial understanding in built environments. By reducing reliance on semantic annotations, the proposed method may improve the scalability and accessibility of indoor localization systems. However, accurate floorplan-based localization could also raise privacy and security concerns if deployed without consent, for example in unauthorized indoor tracking or surveillance-sensitive environments. Practical deployment should therefore require appropriate user consent, access control to floorplan data, and safeguards against misuse in private or restricted spaces.
\clearpage
\newpage
\input{checklist.tex}

\end{document}

%% file: checklist.tex
\section*{NeurIPS Paper Checklist}

\begin{enumerate}

\item {\bf Claims}
    \item[] Question: Do the main claims made in the abstract and introduction accurately reflect the paper's contributions and scope?
    \item[] Answer: \answerYes{}
    \item[] Justification: The abstract and introduction summarize the semantic-free visual floorplan localization setting, the proposed dense-to-ray candidate generation and contrastive disambiguation modules, and the empirical evaluation scope. The claims are supported by the comparative, ablation, and parametric studies in the main paper and appendix.

\item {\bf Limitations}
    \item[] Question: Does the paper discuss the limitations of the work performed by the authors?
    \item[] Answer: \answerYes{}
    \item[] Justification: The paper includes a dedicated Limitations paragraph discussing the two-stage nature of the framework, the dependency between candidate generation and reranking, and the additional inference latency.

\item {\bf Theory assumptions and proofs}
    \item[] Question: For each theoretical result, does the paper provide the full set of assumptions and a complete (and correct) proof?
    \item[] Answer: \answerNA{}
    \item[] Justification: The paper does not present formal theoretical theorems or proofs; the mathematical formulations are used to describe the proposed model, loss, and inference procedure.

    \item {\bf Experimental result reproducibility}
    \item[] Question: Does the paper fully disclose all the information needed to reproduce the main experimental results of the paper to the extent that it affects the main claims and/or conclusions of the paper (regardless of whether the code and data are provided or not)?
    \item[] Answer: \answerYes{}
    \item[] Justification: The paper specifies the datasets, official splits, evaluation metrics, model components, training settings, hyperparameters, and inference procedure in the main paper and appendix.

\item {\bf Open access to data and code}
    \item[] Question: Does the paper provide open access to the data and code, with sufficient instructions to faithfully reproduce the main experimental results, as described in supplemental material?
    \item[] Answer: \answerYes{}
    \item[] Justification: We provide anonymized code in the supplementary material, including environment setup, data preparation instructions, and scripts/commands for reproducing the main experimental results. The experiments are conducted on publicly available datasets, and detailed instructions for accessing and preprocessing the data are included.

\item {\bf Experimental setting/details}
    \item[] Question: Does the paper specify all the training and test details (e.g., data splits, hyperparameters, how they were chosen, type of optimizer) necessary to understand the results?
    \item[] Answer: \answerYes{}
    \item[] Justification: The experimental setup describes the datasets, data splits, FOV settings, floorplan resolutions, metrics, baselines, optimizer, learning rates, batch sizes, training epochs, sampling parameters, and inference hyperparameters.

\item {\bf Experiment statistical significance}
    \item[] Question: Does the paper report error bars suitably and correctly defined or other appropriate information about the statistical significance of the experiments?
    \item[] Answer: \answerNo{}
    \item[] Justification: The paper reports deterministic recall metrics following existing visual floorplan localization protocols, but does not report error bars or confidence intervals due to the high computational cost of repeated full training and evaluation runs.

\item {\bf Experiments compute resources}
    \item[] Question: For each experiment, does the paper provide sufficient information on the computer resources (type of compute workers, memory, time of execution) needed to reproduce the experiments?
    \item[] Answer: \answerYes{}
    \item[] Justification: The appendix reports the GPU type used for training and provides single-query inference time comparisons. The training details specify the optimization settings and training epochs for the proposed modules.
    
\item {\bf Code of ethics}
    \item[] Question: Does the research conducted in the paper conform, in every respect, with the NeurIPS Code of Ethics \url{https://neurips.cc/public/EthicsGuidelines}?
    \item[] Answer: \answerYes{}
    \item[] Justification: The work uses public indoor scene datasets and focuses on semantic-free floorplan localization. We are not aware of any deviation from the NeurIPS Code of Ethics.

\item {\bf Broader impacts}
    \item[] Question: Does the paper discuss both potential positive societal impacts and negative societal impacts of the work performed?
    \item[] Answer: \answerYes{}
    \item[] Justification: The paper discusses potential positive impacts in indoor robotics, AR, assistive navigation, and scalable indoor localization, as well as possible privacy and security risks such as unauthorized indoor tracking or misuse in surveillance-sensitive environments.
    
\item {\bf Safeguards}
    \item[] Question: Does the paper describe safeguards that have been put in place for responsible release of data or models that have a high risk for misuse (e.g., pre-trained language models, image generators, or scraped datasets)?
    \item[] Answer: \answerNA{}
    \item[] Justification: The paper does not release high-risk generative models, scraped datasets, or dual-use pretrained models. It uses existing indoor localization datasets and proposes an algorithmic framework.

\item {\bf Licenses for existing assets}
    \item[] Question: Are the creators or original owners of assets (e.g., code, data, models), used in the paper, properly credited and are the license and terms of use explicitly mentioned and properly respected?
    \item[] Answer: \answerNo{}
    \item[] Justification: The paper cites the datasets, baselines, and pretrained models used in the experiments, but does not explicitly list all asset licenses and terms of use. We will add complete license information in the final version.

\item {\bf New assets}
    \item[] Question: Are new assets introduced in the paper well documented and is the documentation provided alongside the assets?
    \item[] Answer: \answerNA{}
    \item[] Justification: The paper does not introduce a new dataset or benchmark asset. The contribution is a localization method evaluated on existing datasets.

\item {\bf Crowdsourcing and research with human subjects}
    \item[] Question: For crowdsourcing experiments and research with human subjects, does the paper include the full text of instructions given to participants and screenshots, if applicable, as well as details about compensation (if any)?
    \item[] Answer: \answerNA{}
    \item[] Justification: The paper does not involve crowdsourcing experiments or research with human subjects.

\item {\bf Institutional review board (IRB) approvals or equivalent for research with human subjects}
    \item[] Question: Does the paper describe potential risks incurred by study participants, whether such risks were disclosed to the subjects, and whether Institutional Review Board (IRB) approvals (or an equivalent approval/review based on the requirements of your country or institution) were obtained?
    \item[] Answer: \answerNA{}
    \item[] Justification: The paper does not involve human subjects, crowdsourcing, or user studies.

\item {\bf Declaration of LLM usage}
    \item[] Question: Does the paper describe the usage of LLMs if it is an important, original, or non-standard component of the core methods in this research? Note that if the LLM is used only for writing, editing, or formatting purposes and does \emph{not} impact the core methodology, scientific rigor, or originality of the research, declaration is not required.
    \item[] Answer: \answerNA{}
    \item[] Justification: The core method development and experiments do not involve LLMs as an important, original, or non-standard component. Any language editing or formatting assistance does not affect the scientific methodology or results.

\end{enumerate}

%% file: neurips_2026.bib
@inproceedings{karkus2018particle,
  title={Particle filter networks with application to visual localization},
  author={Karkus, Peter and Hsu, David and Lee, Wee Sun},
  booktitle={Conference on robot learning},
  pages={169--178},
  year={2018},
  organization={PMLR}
}

@inproceedings{min2022laser,
  title={Laser: Latent space rendering for 2d visual localization},
  author={Min, Zhixiang and Khosravan, Naji and Bessinger, Zachary and Narayana, Manjunath and Kang, Sing Bing and Dunn, Enrique and Boyadzhiev, Ivaylo},
  booktitle={Proceedings of the IEEE/CVF Conference on Computer Vision and Pattern Recognition},
  pages={11122--11131},
  year={2022}
}

@inproceedings{chen2024f3loc,
  title={F3Loc: Fusion and Filtering for Floorplan Localization},
  author={Chen, Changan and Wang, Rui and Vogel, Christoph and Pollefeys, Marc},
  booktitle={Proceedings of the IEEE/CVF Conference on Computer Vision and Pattern Recognition},
  pages={18029--18038},
  year={2024}
}

@inproceedings{zheng2020structured3d,
  title={Structured3d: A large photo-realistic dataset for structured 3d modeling},
  author={Zheng, Jia and Zhang, Junfei and Li, Jing and Tang, Rui and Gao, Shenghua and Zhou, Zihan},
  booktitle={Computer Vision--ECCV 2020: 16th European Conference, Glasgow, UK, August 23--28, 2020, Proceedings, Part IX 16},
  pages={519--535},
  year={2020},
  organization={Springer}
}

@inproceedings{xia2018gibson,
  title={Gibson env: Real-world perception for embodied agents},
  author={Xia, Fei and Zamir, Amir R and He, Zhiyang and Sax, Alexander and Malik, Jitendra and Savarese, Silvio},
  booktitle={Proceedings of the IEEE conference on computer vision and pattern recognition},
  pages={9068--9079},
  year={2018}
}

@article{2017NetVLAD,
  title={NetVLAD: CNN architecture for weakly supervised place recognition},
  author={ Arandjelovic, Relja  and  Gronat, Petr  and  Torii, Akihiko  and  Pajdla, Tomas  and  Sivic, Josef },
  journal={IEEE Transactions on Pattern Analysis \& Machine Intelligence},
  pages={1-1},
  year={2017},
}

@inproceedings{sarlin2019coarse, 
  title={From coarse to fine: Robust hierarchical localization at large scale},
  author={Sarlin, Paul-Edouard and Cadena, Cesar and Siegwart, Roland and Dymczyk, Marcin},
  booktitle={Proceedings of the IEEE/CVF conference on computer vision and pattern recognition},
  pages={12716--12725},
  year={2019}
}

@article{sattler2016efficient,
  title={Efficient \& effective prioritized matching for large-scale image-based localization},
  author={Sattler, Torsten and Leibe, Bastian and Kobbelt, Leif},
  journal={IEEE transactions on pattern analysis and machine intelligence},
  volume={39},
  number={9},
  pages={1744--1756},
  year={2016},
  publisher={IEEE}
}

@article{li2024flona,
  title={FloNa: Floor Plan Guided Embodied Visual Navigation},
  author={Li, Jiaxin and Huang, Weiqi and Wang, Zan and Liang, Wei and Di, Huijun and Liu, Feng},
  journal={arXiv preprint arXiv:2412.18335},
  year={2024}
}

@inproceedings{chu2015you, 
  title={You are here: Mimicking the human thinking process in reading floor-plans},
  author={Chu, Hang and Kim, Dong Ki and Chen, Tsuhan},
  booktitle={Proceedings of the IEEE International Conference on Computer Vision},
  pages={2210--2218},
  year={2015}
}

@inproceedings{boniardi2019robot, 
  title={Robot localization in floor plans using a room layout edge extraction network},
  author={Boniardi, Federico and Valada, Abhinav and Mohan, Rohit and Caselitz, Tim and Burgard, Wolfram},
  booktitle={2019 IEEE/RSJ International Conference on Intelligent Robots and Systems (IROS)},
  pages={5291--5297},
  year={2019},
  organization={IEEE}
}

@inproceedings{howard2022lalaloc++, 
  title={LaLaLoc++: Global floor plan comprehension for layout localisation in unvisited environments},
  author={Howard-Jenkins, Henry and Prisacariu, Victor Adrian},
  booktitle={European Conference on Computer Vision},
  pages={693--709},
  year={2022},
  organization={Springer}
}

@inproceedings{howard2021lalaloc, 
  title={Lalaloc: Latent layout localisation in dynamic, unvisited environments},
  author={Howard-Jenkins, Henry and Ruiz-Sarmiento, Jose-Raul and Prisacariu, Victor Adrian},
  booktitle={Proceedings of the IEEE/CVF International Conference on Computer Vision},
  pages={10107--10116},
  year={2021}
}

@inproceedings{he2016deep,
  title={Deep residual learning for image recognition},
  author={He, Kaiming and Zhang, Xiangyu and Ren, Shaoqing and Sun, Jian},
  booktitle={Proceedings of the IEEE conference on computer vision and pattern recognition},
  pages={770--778},
  year={2016}
}

@article{jonschkowski2016end,
  title={End-to-end learnable histogram filters},
  author={Jonschkowski, Rico and Brock, Oliver},
  year={2016}
}

@article{vaswani2017attention,
  title={Attention is all you need},
  author={Vaswani, Ashish and Shazeer, Noam and Parmar, Niki and Uszkoreit, Jakob and Jones, Llion and Gomez, Aidan N and Kaiser, {\L}ukasz and Polosukhin, Illia},
  journal={Advances in neural information processing systems},
  volume={30},
  year={2017}
}

@inproceedings{balntas2018relocnet,
  title={Relocnet: Continuous metric learning relocalisation using neural nets},
  author={Balntas, Vassileios and Li, Shuda and Prisacariu, Victor},
  booktitle={Proceedings of the European conference on computer vision (ECCV)},
  pages={751--767},
  year={2018}
}

@inproceedings{brachmann2017dsac,
  title={Dsac-differentiable ransac for camera localization},
  author={Brachmann, Eric and Krull, Alexander and Nowozin, Sebastian and Shotton, Jamie and Michel, Frank and Gumhold, Stefan and Rother, Carsten},
  booktitle={Proceedings of the IEEE conference on computer vision and pattern recognition},
  pages={6684--6692},
  year={2017}
}

@article{kingma2014adam,
  title={Adam: A method for stochastic optimization},
  author={Kingma, Diederik P},
  journal={arXiv preprint arXiv:1412.6980},
  year={2014}
}

@inproceedings{dellaert1999monte, 
  title={Monte carlo localization for mobile robots},
  author={Dellaert, Frank and Fox, Dieter and Burgard, Wolfram and Thrun, Sebastian},
  booktitle={Proceedings 1999 IEEE international conference on robotics and automation (Cat. No. 99CH36288C)},
  volume={2},
  pages={1322--1328},
  year={1999},
  organization={IEEE}
}

@inproceedings{mendez2018sedar, 
  title={Sedar-semantic detection and ranging: Humans can localise without lidar, can robots?},
  author={Mendez, Oscar and Hadfield, Simon and Pugeault, Nicolas and Bowden, Richard},
  booktitle={2018 IEEE International Conference on Robotics and Automation (ICRA)},
  pages={6053--6060},
  year={2018},
  organization={IEEE}
}

@inproceedings{winterhalter2015accurate,
  title={Accurate indoor localization for RGB-D smartphones and tablets given 2D floor plans},
  author={Winterhalter, Wera and Fleckenstein, Freya and Steder, Bastian and Spinello, Luciano and Burgard, Wolfram},
  booktitle={2015 IEEE/RSJ International Conference on Intelligent Robots and Systems (IROS)},
  pages={3138--3143},
  year={2015},
  organization={IEEE}
}

@article{bishop2001introduction,
  title={An introduction to the kalman filter},
  author={Bishop, Gary and Welch, Greg and others},
  journal={Proc of SIGGRAPH, Course},
  volume={8},
  number={27599-23175},
  pages={41},
  year={2001}
}

@article{mendez2020sedar,
  title={SeDAR: reading floorplans like a human—using deep learning to enable human-inspired localisation},
  author={Mendez, Oscar and Hadfield, Simon and Pugeault, Nicolas and Bowden, Richard},
  journal={International Journal of Computer Vision},
  volume={128},
  number={5},
  pages={1286--1310},
  year={2020},
  publisher={Springer}
}

@article{grader2025supercharging,
  title={Supercharging Floorplan Localization with Semantic Rays},
  author={Grader, Yuval and Averbuch-Elor, Hadar},
  journal={arXiv preprint arXiv:2507.09291},
  year={2025}
}

@inproceedings{chen20253dp,
  title={Perspective from a Higher Dimension: Can 3D Geometric Priors Help Visual Floorplan Localization?},
  author={Chen, Bolei and Kang, Jiaxu and Yang, Haonan and Zhong, Ping and Wang, Jianxin},
  booktitle={Proceedings of the 33nd ACM International Conference on Multimedia},
  year={2025}
}

@inproceedings{liu2017efficient,
  title={Efficient global 2d-3d matching for camera localization in a large-scale 3d map},
  author={Liu, Liu and Li, Hongdong and Dai, Yuchao},
  booktitle={Proceedings of the IEEE International Conference on Computer Vision},
  pages={2372--2381},
  year={2017}
}

@inproceedings{deng2009imagenet,
  title={Imagenet: A large-scale hierarchical image database},
  author={Deng, Jia and Dong, Wei and Socher, Richard and Li, Li-Jia and Li, Kai and Fei-Fei, Li},
  booktitle={2009 IEEE conference on computer vision and pattern recognition},
  pages={248--255},
  year={2009},
  organization={Ieee}
}

@article{yang2024depth,
  title={Depth anything v2},
  author={Yang, Lihe and Kang, Bingyi and Huang, Zilong and Zhao, Zhen and Xu, Xiaogang and Feng, Jiashi and Zhao, Hengshuang},
  journal={Advances in Neural Information Processing Systems},
  volume={37},
  pages={21875--21911},
  year={2024}
}

@article{chen2025perspective,
  title={Perspective from a Broader Context: Can Room Style Knowledge Help Visual Floorplan Localization?},
  author={Chen, Bolei and Yan, Shengsheng and Cui, Yongzheng and Kang, Jiaxu and Zhong, Ping and Wang, Jianxin},
  journal={arXiv preprint arXiv:2508.01216},
  year={2025}
}

@article{huang2025floor,
  title={Floor Plan-Guided Visual Navigation Incorporating Depth and Directional Cues},
  author={Huang, Wei and Li, Jiaxin and Wan, Zang and Di, Huijun and Liang, Wei and Yang, Zhu},
  journal={arXiv preprint arXiv:2511.01493},
  year={2025}
}

@inproceedings{panek2022meshloc,
  title={Meshloc: Mesh-based visual localization},
  author={Panek, Vojtech and Kukelova, Zuzana and Sattler, Torsten},
  booktitle={European Conference on Computer Vision},
  pages={589--609},
  year={2022},
  organization={Springer}
}

@inproceedings{kukelova2008automatic,
  title={Automatic generator of minimal problem solvers},
  author={Kukelova, Zuzana and Bujnak, Martin and Pajdla, Tomas},
  booktitle={European Conference on Computer Vision},
  pages={302--315},
  year={2008},
  organization={Springer}
}

@article{fischler1981random,
  title={Random sample consensus: a paradigm for model fitting with applications to image analysis and automated cartography},
  author={Fischler, Martin A and Bolles, Robert C},
  journal={Communications of the ACM},
  volume={24},
  number={6},
  pages={381--395},
  year={1981},
  publisher={ACM New York, NY, USA}
}

@article{barath2021graph,
  title={Graph-cut RANSAC: Local optimization on spatially coherent structures},
  author={Barath, Daniel and Matas, Jiri},
  journal={IEEE transactions on pattern analysis and machine intelligence},
  volume={44},
  number={9},
  pages={4961--4974},
  year={2021},
  publisher={IEEE}
}

@inproceedings{kendall2017geometric,
  title={Geometric loss functions for camera pose regression with deep learning},
  author={Kendall, Alex and Cipolla, Roberto},
  booktitle={Proceedings of the IEEE conference on computer vision and pattern recognition},
  pages={5974--5983},
  year={2017}
}

@article{oquab2023dinov2,
  title={Dinov2: Learning robust visual features without supervision},
  author={Oquab, Maxime and Darcet, Timoth{\'e}e and Moutakanni, Th{\'e}o and Vo, Huy and Szafraniec, Marc and Khalidov, Vasil and Fernandez, Pierre and Haziza, Daniel and Massa, Francisco and El-Nouby, Alaaeldin and others},
  journal={arXiv preprint arXiv:2304.07193},
  year={2023}
}
